\documentclass[11pt,a4paper]{article}
\usepackage[hyperref]{acl2020}
\usepackage{times}
\usepackage{latexsym}

\usepackage{microtype}

\usepackage{url}
\usepackage{textcomp}

\usepackage{amsmath}
\usepackage{amsfonts}
\usepackage{amssymb}
\usepackage{bm}
\usepackage{color}
\usepackage{multirow}
\usepackage{subfig}
\usepackage{enumitem}
\usepackage{multirow}
\usepackage{makecell}
\usepackage{arydshln}
\usepackage{graphicx}
\usepackage{subfig}
\usepackage{algpseudocode}

\aclfinalcopy

\definecolor{zptu}{RGB}{18, 141, 21}

\definecolor{shuo}{RGB}{250, 0, 0}

\begin{document}

\title{On the Inference Calibration of Neural Machine Translation}

\author{
Shuo Wang$^\star$\thanks{~~Work was done when Shuo Wang was interning at Tencent AI Lab under the Rhino-Bird Elite Training Program.} \quad\quad Zhaopeng Tu$^\top$ \quad\quad Shuming Shi$^\top$ \quad\quad Yang Liu$^\star$$^\dagger$ \\
$^\star$Institute for Artificial Intelligence\\
Department of Computer Science and Technology, Tsinghua University\\
Beijing National Research Center for Information Science and Technology\\
$^\top$Tencent AI Lab\\
$^\dagger$Beijing Academy of Artificial Intelligence \\
Beijing Advanced Innovation Center for Language Resources\\
$^\star${\tt \{wangshuo.thu, liuyang.china\}@gmail.com}\\
$^\top${\tt \{zptu, shumingshi\}@tencent.com}
}

\maketitle

\begin{abstract}
Confidence calibration, which aims to make model predictions equal to the true correctness measures, is important for neural machine translation (NMT) because it is able to offer useful indicators of translation errors in the generated output. % by LY
While prior studies have shown that NMT models trained with label smoothing are well-calibrated on the ground-truth training data, we find that miscalibration still remains a severe challenge for NMT during inference due to the discrepancy between training and inference. % by LY
By carefully designing experiments on three language pairs, our work provides in-depth analyses of the correlation between calibration and translation performance as well as linguistic properties of miscalibration and reports a number of interesting findings that might help humans better analyze, understand and improve NMT models. % by LY
Based on these observations, we further propose a new graduated label smoothing method that can improve both inference calibration and translation performance.~\footnote{The source code is available at \url{https://github.com/shuo-git/InfECE}.}
% by LY
\end{abstract}

\section{Introduction}

Calibration requires that the probability a model assigns to a prediction (i.e., {\em confidence}) equals to the correctness measure of the prediction (i.e., {\em accuracy}). 
Calibrated models are important in user-facing applications such as natural language processing~\cite{Nguyen:2015:EMNLP} and speech recognition~\cite{Yu:2011:TASLP}, in which one needs to assess the confidence of a prediction.
For example, in computer-assisted translation, a calibrated machine translation model is able to tell a user when the model's predictions are likely to be incorrect, which is helpful for the user to correct errors.
% and improve the quality of translation.

The study of calibration on classification tasks has a long history, from  statistical machine learning~\cite{Platt:1999:ALMC,Niculescu:2005:ICML} to  deep learning~\cite{Guo:2017:ICML}. However, calibration on structured generation tasks such as neural machine translation (NMT) has not been well studied.
Recently,~\newcite{Muller:2019:arXiv} and ~\newcite{Kumar:2019:ICLR-Workshop} studied the calibration of NMT in the training setting, and found that NMT trained with label smoothing~\cite{Szegedy:2016:CVPR} is well-calibrated.
We believe that this setting would cover up a central problem of NMT, {\em the exposure bias}~\cite{Ranzato:2015:arXiv} -- the training-inference discrepancy caused by teacher forcing in the training of auto-regressive models.

\begin{figure}[t]
    \centering
    \subfloat[Training]{
    \includegraphics[height=0.236\textwidth]{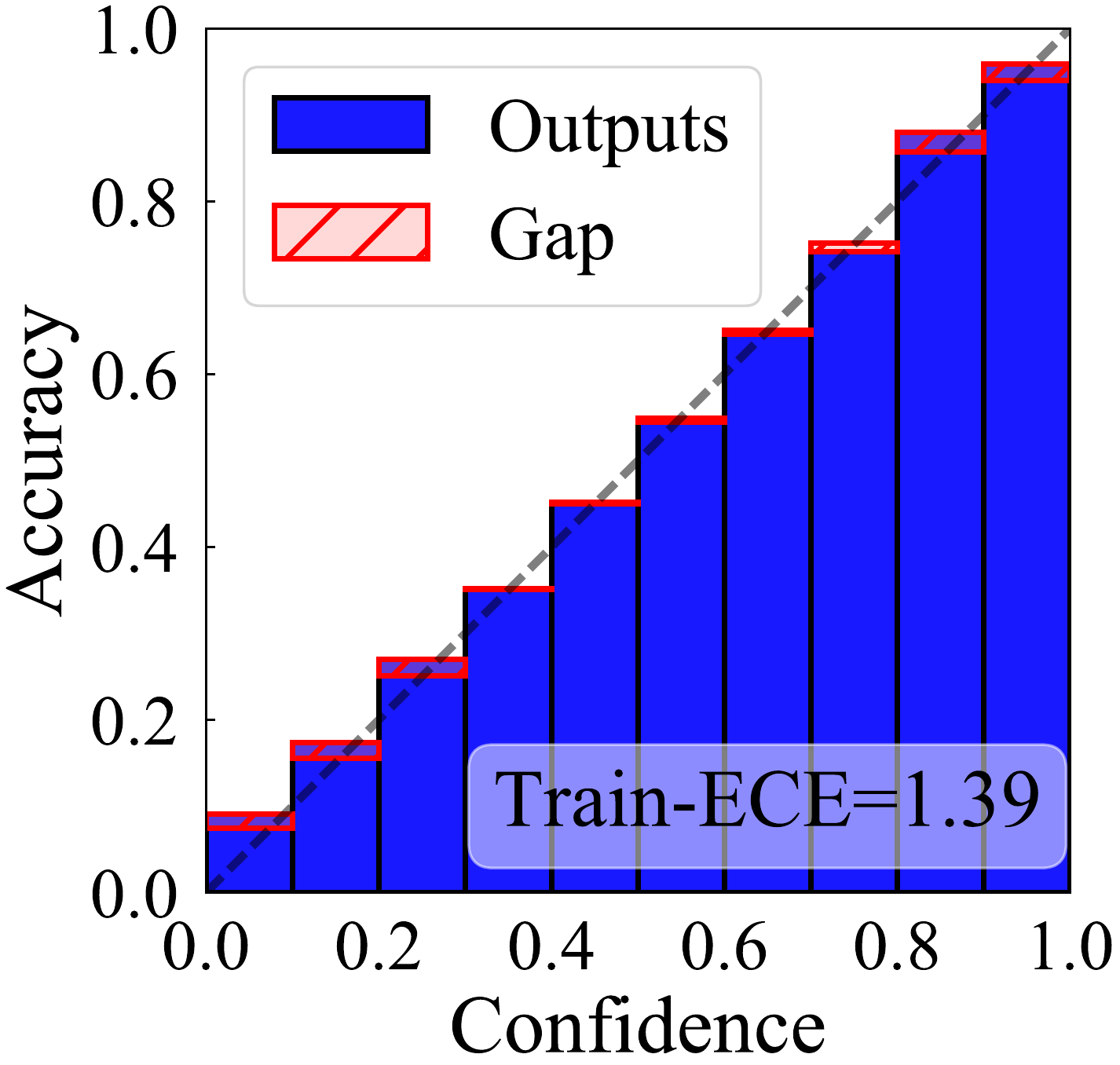}}
    \hspace{0.0\textwidth}
    \subfloat[Inference]{
    \includegraphics[height=0.236\textwidth]{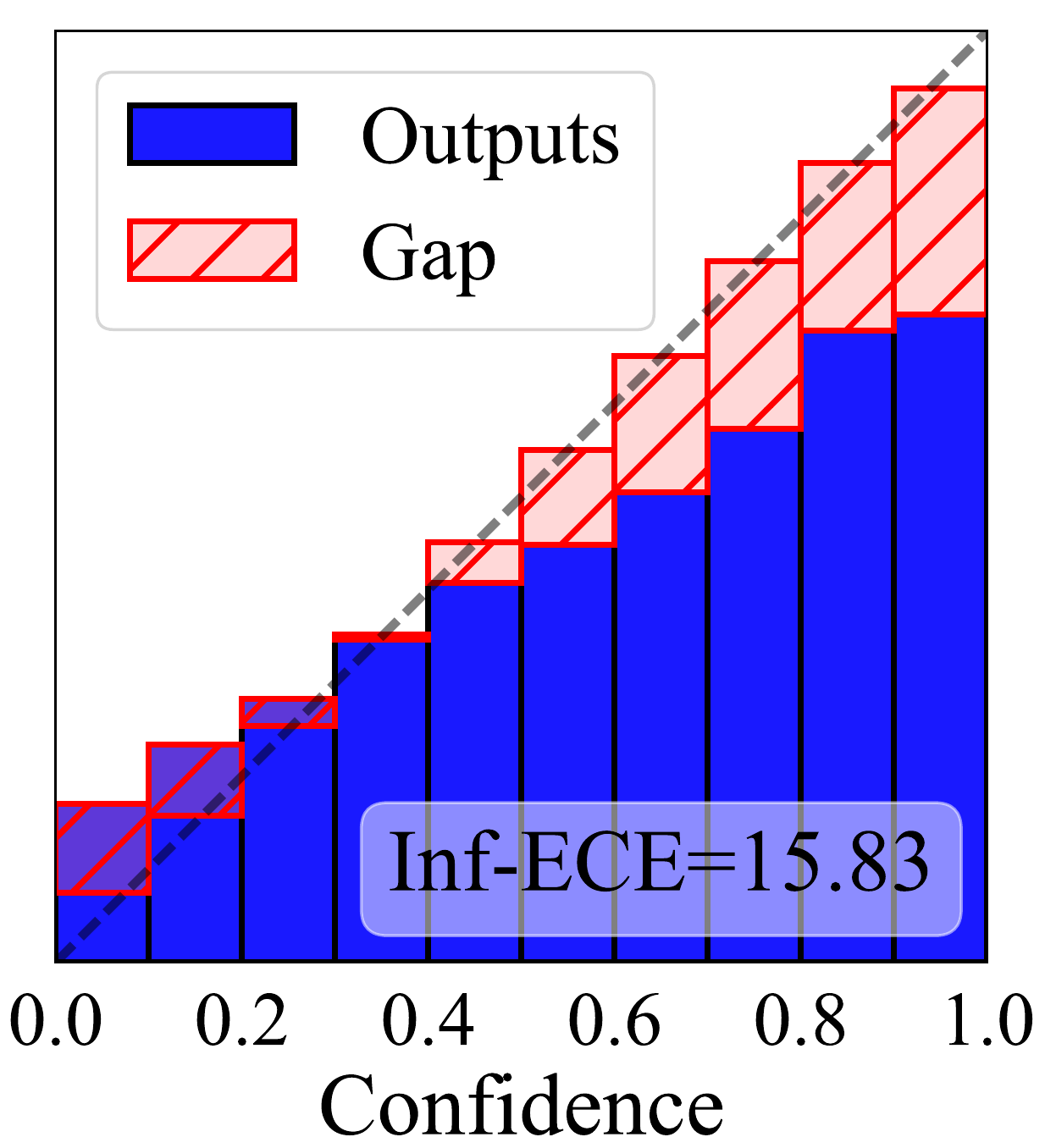}} \hspace{0.02\textwidth}
\caption{Reliability diagrams in training and inference for the WMT14 En-De task. ``Gap'' denotes the difference between confidence and accuracy. Smaller gaps denotes better calibrated outputs. 
%Gaps above the dashed line denote under-estimation, and those below the dashed line denote over-estimation.
We find that the average gaps between confidence and accuracy are much larger in inference than in training (i.e., 15.83 $>$ 1.39).
}
\label{fig:ende-diag}
\end{figure}

In response to this problem, this work focuses on the calibration of NMT in inference, which can better reflect the generative capacity of NMT models. To this end, we use translation error rate (TER)~\cite{Snover:2006:MT} to automatically annotate the correctness of generated tokens, which makes it feasible to evaluate calibration in inference.
Experimental results on several datasets across language pairs show that even trained with label smoothing, NMT models still suffer from miscalibration errors in inference. Figure~\ref{fig:ende-diag} shows an example.
% , \shuo{which is visualized in Figure~\ref{fig:ende-diag}}.
While modern neural networks on classification tasks have been found to be miscalibrated in the direction of {\em over-estimation} (i.e., confidence $>$ accuracy)~\cite{Guo:2017:ICML}, NMT models are also {\em under-estimated} (i.e., confidence $<$ accuracy) on low-confidence predictions.
In addition, we found that miscalibrated predictions correlate well with the translation errors in inference.
Specifically, the over-estimated predictions correlate more with over-translation and mis-translation errors, while the under-estimated predictions correlate more with under-translation errors.
This demonstrates the necessity of studying inference calibration for NMT.

By investigating the linguistic properties of miscalibrated tokens in NMT outputs, we have several interesting findings:
\begin{itemize}
    \item {\em Frequency}: Low-frequency tokens generally suffer from under-estimation. Moreover, low-frequency tokens contribute more to over-estimation than high-frequency tokens, especially on large-scale data.
    \item {\em Position}: Over-estimation does not have a bias on the position of generated tokens, while under-estimation occurs more in the left part of a generated sentence than in the right part.
    \item {\em Fertility}: Predicted tokens that align to more than one source token (``fertility$\ge$2'') suffer more from under-estimation, while tokens with fertility $<1$ suffer from over-estimation.
    \item {\em Syntactic Roles}: Content tokens are more likely to suffer from miscalibration than content-free tokens. Specifically, verbs are more likely to suffer from over-estimation than under-estimation.
    \item {{\em Word Granularity}}: sub-words suffer more from both over-estimation and under-estimation, while full words are less likely to be miscalibrated.
\end{itemize}

Inspired by the finding that miscalibration on classification tasks is closely related to lack of regularization and increased model size~\cite{Guo:2017:ICML}, we revisit these techniques on the NMT (i.e., structured generation) task:
\begin{itemize}
    \item {\em Regularization Techniques}: We investigate label smoothing and dropout~\cite{Hinton:2012:arXiv}, which directly affect the confidence estimation. Both label smoothing and dropout improve the inference calibration by alleviating the over-estimation. Label smoothing is the key for well-calibration, which is essential for maintaining translation performance for inference in large search space. Inspired by this finding, we propose a novel {\em graduated label smoothing} approach, in which the smoothing penalty for high-confidence predictions is higher than that for low-confidence predictions. The graduated label smoothing can improve translation performance by alleviating inference miscalibration.
    \item {\em Model Size}: Increasing model size consistently improves translation performance at the cost of negatively affecting inference calibration. The problem can be alleviated by increasing the capacity of encoder only, which  maintains the inference calibration and obtains a further improvement of translation performance in large search space.
\end{itemize}

To summarize, the main contributions of our work are listed as follows:
\begin{itemize}
    \item We demonstrate the necessity of studying inference calibration for NMT, which can serve as useful indicators of translation errors.
    \item We reveal certain linguistic properties of miscalibrated predictions in NMT, which provides potentially useful information for the design of training procedures.
    \item We revisit recent advances in architectures and regularization techniques, and provide variants that can boost translation performance by improving inference calibration.
\end{itemize}

\begin{figure*}[ht]
    \centering
    \includegraphics[width=0.8\textwidth]{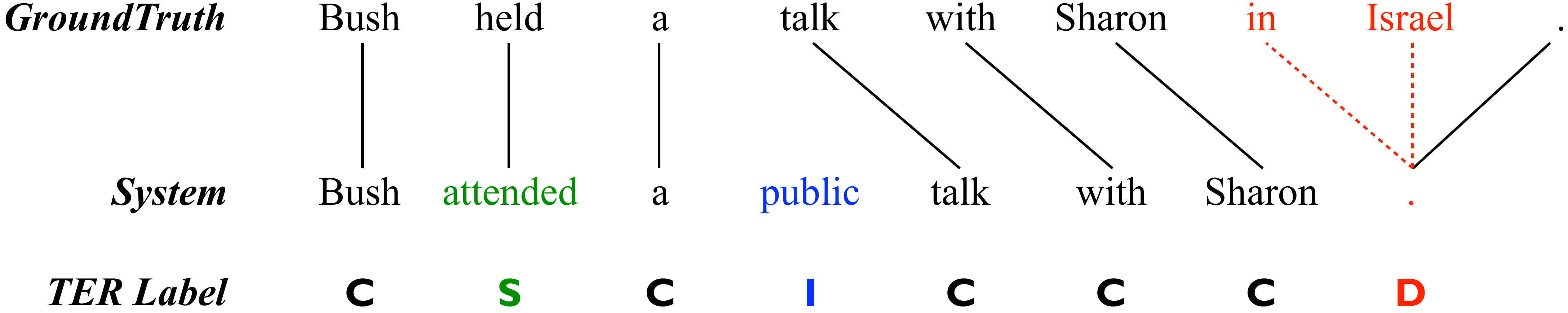}
\caption{An example of TER labels. ``C'': correct, ``S'': substitution, corresponding to mis-translation, ``I'': insertion, corresponding to over-translation, ``D'': deletion, corresponding to under-translation. Dash line denotes mapping the label ``D'' from the ground-truth sequence to the generated sequence.}
\label{fig:ter_exp}
\end{figure*}

\section{Related Work}

\paragraph{Calibration on Classification}
Calibration on classification tasks has been studied for a long history in the statistics literature, including Platt scaling~\cite{Platt:1999:ALMC}, isotonic regression~\cite{Niculescu:2005:ICML} and many other methods for non-binary classification~\cite{Zadrozny:2002:KDD,Menon:2012:ICML,Zhong:2013:IJCAI}.
% \citet{Niculescu:2005:ICML} evaluated a number of classical models systematically and found models trained on conditional likelihood to be well-calibrated while naive Bayes and SVMs were poorly calibrated. 
For modern deep neural networks, \citet{Guo:2017:ICML} demonstrated that recent advances in training and model architecture have strong effects on the calibration. \citet{Szegedy:2016:CVPR} propose the label smoothing technique which can effectively reduce the calibration error. \citet{Ding:2019:arXiv} extend label smoothing to adaptive label regularization.

\paragraph{Calibration on Structured Prediction} Different from classification tasks, most natural language processing (NLP) tasks deal with complex structures~\cite{Kuleshov:2015:NeurIPS}.  \citet{Nguyen:2015:EMNLP} verified the finding of \citet{Niculescu:2005:ICML} in NLP tasks on log-linear structured models. For NMT, some works directed their attention to the uncertainty in prediction~\cite{Ott:2018:ICML,Wang:2019:EMNLP}, \citet{Kumar:2019:ICLR-Workshop} studied the calibration of several NMT models and found that the end of a sentence is severely miscalibrated. \citet{Muller:2019:arXiv} investigated the effect of label smoothing, finding that NMT models are well-calibrated in training. Different from previous works, we are interested in the calibration of NMT models in inference, given that the training and inference are discrepant for standard NMT models~\cite{Vaswani:2017:NeurIPS}.

\section{Definitions of Calibration}

\subsection{Neural Machine Translation}

\paragraph{Training}
In machine translation task, an NMT model $F$: $\textbf{x} \rightarrow \textbf{y}$ maximizes the probability of a target sequence $\textbf{y} = \{y_1,...,y_T\}$ given a source sentence $\textbf{x} = \{x_1,...,x_S\}$:
\begin{equation}
    P(\textbf{y}|\textbf{x};\bm{\theta}) = \displaystyle \prod_{t=1}^{T} P(y_t|\textbf{y}_{<t},\textbf{x};\bm{\theta}),
    \label{eqn:mle}
\end{equation}
where $\bm{\theta}$ is a set of model parameters and $\textbf{y}_{<t}$ is a partial translation. At each time step, the model generates an output token of the highest probability based on the source sentence $\textbf{x}$ and the partial translation $\textbf{y}_{<t}$.
The training objective is to minimize the negative log-likelihood loss on the training corpus. 

\paragraph{Inference}
NMT models are trained on the ground-truth data distribution ({\em teaching forcing}), while in inference the models generate target tokens based on previous model predictions, which can be erroneous. 
The training-inference discrepancy caused by teacher forcing in maximum likelihood estimation training (Equation~\ref{eqn:mle}) is often referred to as {\em exposure bias}~\cite{Ranzato:2015:arXiv}.
% In practice, beam search is employed to decode output sequences, which limits the search space by considering only a fixed number of hypotheses.
In this work, we aim to investigate the calibration of NMT in inference, which we believe can better reflect the generation capacity of NMT models.

% \paragraph{Exposure Bias} NMT models are trained on the groundtruth data distribution ({\em teaching forcing}), but in inference the models generate target tokens based on previous model predictions ({\em sampling}), which can be erroneous;

\begin{figure*}[t]
    \centering
    \subfloat[En-Jp]{
    \includegraphics[height=0.236\textwidth]{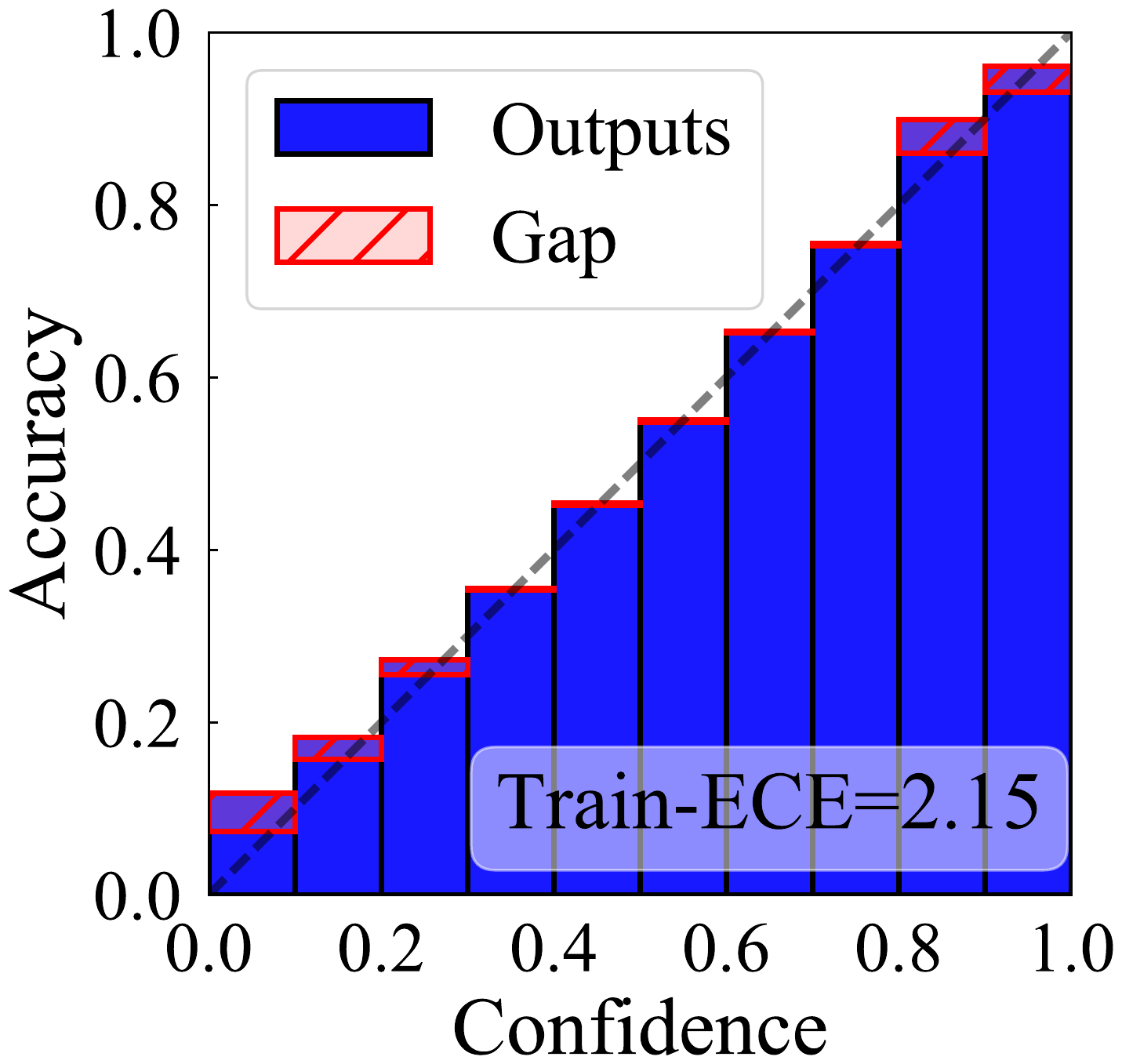} 
    \includegraphics[height=0.236\textwidth]{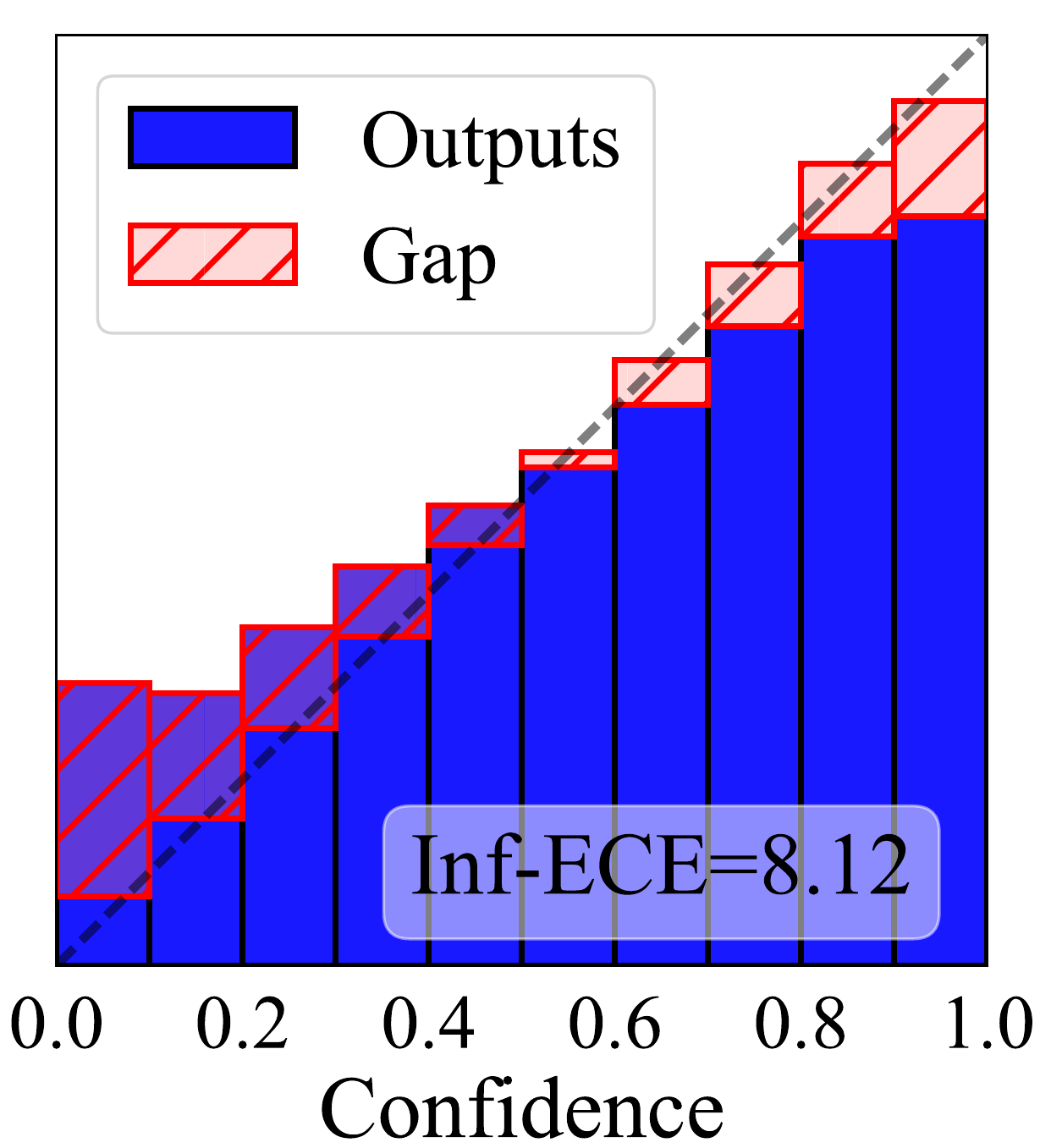}
    } 
    \hspace{0.02\textwidth}
    \subfloat[Zh-En]{
    \includegraphics[height=0.236\textwidth]{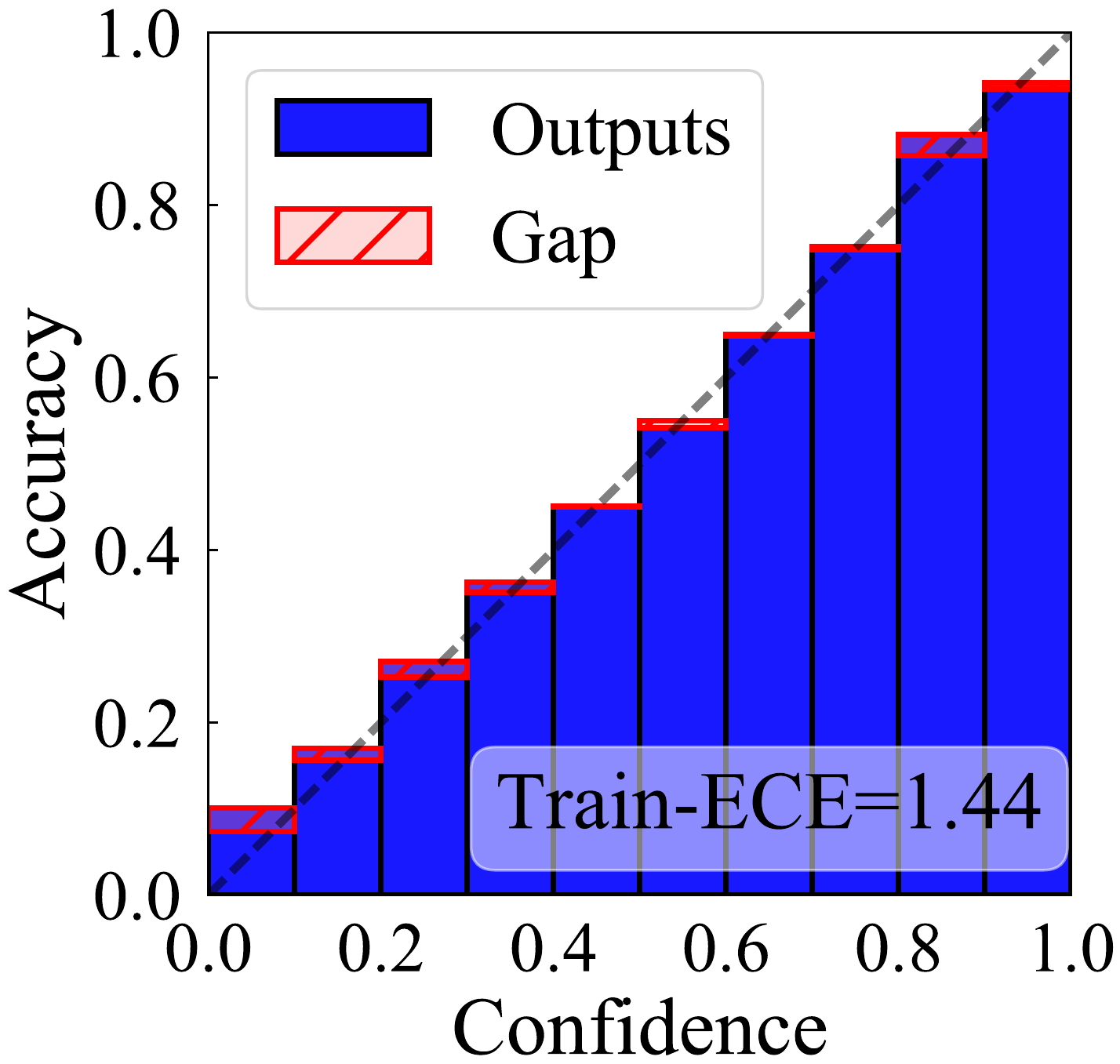} 
    \includegraphics[height=0.236\textwidth]{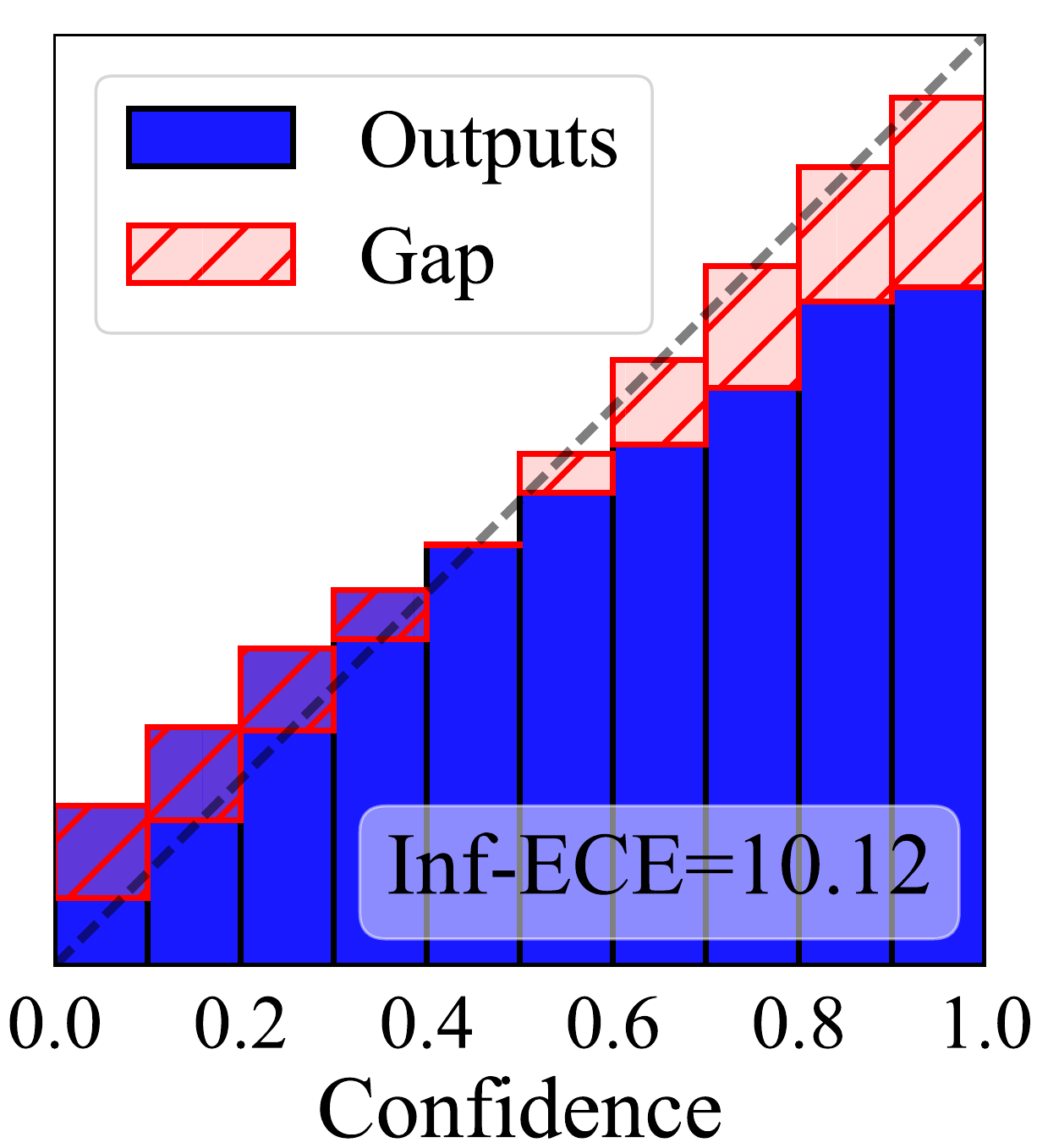}
    }
    
\iffalse
    \centering
    \subfloat[En-Jp]{
    \includegraphics[width=0.27\textwidth]{figures/enjp_train_bar.pdf}} \hspace{0.02\textwidth}
    \subfloat[En-De]{
    \includegraphics[width=0.27\textwidth]{figures/ende_train_bar.pdf}} \hspace{0.02\textwidth}
    \subfloat[Zh-En]{
    \includegraphics[width=0.27\textwidth]{figures/zhen_train_bar.pdf}}
    \\
    \subfloat[En-Jp]{
    \includegraphics[width=0.27\textwidth]{figures/enjp_inf_bar.pdf}} \hspace{0.02\textwidth}
    \subfloat[En-De]{
    \includegraphics[width=0.27\textwidth]{figures/ende_inf_bar.pdf}} \hspace{0.02\textwidth}
    \subfloat[Zh-En]{
    \includegraphics[width=0.27\textwidth]{figures/zhen_inf_bar.pdf}}
\fi
\caption{Reliability diagrams on (a) En-Jp and (b) Zh-En datasets. Left: training, right: inference.}
% Left panel is training calibration and right panel is inference calibration. ``Gap'' denotes the difference between confidence and accuracy, and smaller gaps denote better calibrated outputs. Gaps above the dashed line denote {\em under-estimation}, and the others are {\em over-estimation}.}
\label{fig:acc_conf_bar}
\end{figure*}

\subsection{Calibration of NMT}
\label{sec:cali-nmt}

Calibration requires that the probability a model assigns to a prediction (i.e., {\em confidence}) equals to the true correctness measure of the prediction (i.e., {\em accuracy}).
Modern neural networks have been found to be miscalibrated in the direction of over-estimation~\cite{Guo:2017:ICML}. In this study, we revisit the calibration problem in NMT. If an NMT model is well-calibrated, the gap between the confidence of the generated tokens and the accuracy of them will be small.~\footnote{For example, given 100 predictions, each with confidence 0.7. If the accuracy is also 0.7 (i.e., 70 of the 100 tokens are correct), then the NMT model is well calibrated.}

\paragraph{Expected Calibration Error (ECE)}
ECE is a commonly-used metric to evaluate the miscalibration, which measures the difference in expectation between confidence and accuracy~\cite{Naeini:2015:AAAI}. Specifically, ECE partitions predictions into $M$ bins $\{B_1, \dots, B_M\}$ according to their confidence and takes a weighted average of the bin's accuracy/confidence difference:
\begin{equation}
\label{eq:ece}
    ECE = \sum_{m=1}^M \frac{|B_m|}{N} \Big| acc(B_m) - conf({B_m}) \Big|,
\end{equation}
where $N$ is the number of prediction samples and $|B_m|$ is the number of samples in the $m$-th bin.

\paragraph{ECE in Training and Inference}
In the case of considering just the topmost token in structured prediction tasks (e.g., machine translation), the prediction is $\hat{y} = \arg\max_{y \in \mathcal{V}} P(y)$ with $P(\hat{y})$ as {\em confidence}. The {\em accuracy} $C(\hat{y}) \in \{1, 0\}$ denotes whether the prediction $\hat{y}$ is correct.

In training, the correctness of the prediction $\hat{y}$ is calculated as whether $\hat{y}$ matches the ground-truth token $y_n$: $C(\hat{y}) \in \{1, 0\}$. 
However,  in inference it is not straightforward to measure the accuracy of $\hat{y}$, since it requires to build an alignment between the generated tokens and the ground-truth tokens.

To this end, we turn to the metric of Translation Error Rate (TER)~\cite{Snover:2006:MT}, which measures the number of edits required to change a model output into the ground-truth sequence. Specifically, it assigns a label $l \in \{$C, S, I$\}$ to each generated token. Figure \ref{fig:ter_exp} shows an example of TER labels of each generated token with respect to the reference.
As a side product, TER annotations provide the information of translation errors. 
% \zptu{need to polish this sentence and describe the advantage of listing translation errors}. 
While TER only labels the mis-translation (``S'') and over-translation (``I'') errors, we describe a simple heuristic method to annotate the under-translation error by mapping the label ``D'' from the ground-truth sequence to the generated sequence.

\section{Miscalibration in NMT}

% All tokens are divided into several bins according to their prediction confidence. Each blue rectangle (i.e., Outputs) denotes the average accuracy of the tokens in a bin, and each 
%The dashed line is a reference line, representing well-calibrated outputs, in which case the gap is be zero. 
%In training, output tokens are conditioned on ground-truth prefixes while in inference output tokens are conditioned on machine translation prefixes. We find that the area of gaps in inference is much bigger than in training, indicating that NMT models suffer from miscalibration during inference.}

\paragraph{Data and Setup}
We carried out experiments on three different language pairs, including WAT17 English-Japanese (En-Jp), WMT14 English-German (En-De), and WMT17 Chinese-English (Zh-En). The training datasets consist of 1.9M, 4.5M, and 20.6M sentence pairs respectively.
We employed Byte pair encoding (BPE) \cite{Sennrich:2016:ACL} with 32K merge operations for all the three language pairs.
We used BLEU~\cite{Papineni:2001:ACL} to evaluate the NMT models.
We used the TER toolkit~\cite{Snover:2006:MT} to label whether the tokens in NMT outputs are correctly translated. Normalization was not used, and the maximum shift distance was set to 50.

% The training sets of these three language pairs are of different scales. 
%For the En-Jp task, we followed \citet{morishita:2017:WAT} to use the first two sessions of the English-Japanese dataset in WAT17. The training set consists of 1.9M sentence pairs. For the EN-De task, we used the WMT14 English-German dataset, in which the training set contains 4.5M sentence pairs. For the Zh-En task, we used WMT17 Chinese-English dataset, where the training set consists of 20.6M sentence pairs. 
%We employ Byte pair encoding (BPE) \cite{Sennrich:2016:ACL} with 32K merge operations for all the three language pairs.
%Byte pair encoding (BPE) \cite{Sennrich:2016:ACL} are applied to all the three language pairs to split the words into sub-word units. The merge operations of all the three language pairs are set to 32K. When training NMT models, sentence pairs are batched by approximate length, each batch has roughly 25000 tokens. Only the Zh-En dataset is lowercased.

% which is calculated by the \verb|multi-bleu.perl| script. 
% For the En-Jp and EN-De tasks, we reported the case-sensitive BLEU, and for the Zh-En task, we reported the case-insensitive BLEU. 

The NMT model that we used in our experiments is Transformer~\cite{Vaswani:2017:NeurIPS}. We used base model as default, which consists of a 6-layer encoder and a 6-layer decoder and the hidden size is 512. The model parameters are optimized by Adam~\cite{Kingma:2015:CoRR}, with $\beta_{1}=0.9$, $\beta_{2}=0.98$ and $\epsilon=10^{-9}$. We used the same warm-up strategy for learning rate as \citet{Vaswani:2017:NeurIPS} with $\rm warmup\_steps=4,000$. 
% All experiments were conducted on 8 NVIDIA GTX 1080Ti GPUs.
% \zptu{fill the details here} 

\subsection{Observing Miscalibration}

% \zptu{the diagrams of training and inference ECE should be listed here.}

Reliability diagrams are a visual representation of model calibration, which plot accuracy as a function of confidence~\cite{Niculescu:2005:ICML}.
Specifically, it partitions the output tokens into several bins according to their prediction confidence, and calculate the average confidence and accuracy of each bin.
Figure~\ref{fig:ende-diag} shows the reliability diagrams of both training and inference on En-De and Figure~\ref{fig:acc_conf_bar} shows those on En-Jp and Zh-En. Results are reported on the validation sets. 

{\em NMT still suffers from miscalibration.}
The difference between training and inference ECEs is that when estimating training ECE, NMT models are fed with ground-truth prefixes~\cite{Kumar:2019:ICLR-Workshop,Muller:2019:arXiv}, while for inference ECE, NMT models are fed with previous model predictions.
As seen, the training ECE is very small, indicating that NMT models are well-calibrated in training. This is consistent with the findings of \citet{Kumar:2019:ICLR-Workshop,Muller:2019:arXiv}. 
However, the inference ECE is much higher, suggesting that NMT models still suffer from miscalibration in inference. 

{\em NMT models are miscalibrated in directions of both over- and under-estimation.}
Modern neural networks have been found to be miscalibrated on classification tasks in the direction of over-estimation~\cite{Guo:2017:ICML}. In contrast, NMT models also suffer from under-estimation problems. 
The under-estimation problem is more serious on En-Jp than on Zh-En, which we attribute to the smaller size of the training data of the En-Jp task.
% We also realize that NMT models are negatively affected by not only over-estimation but also under-estimation, different from classification neural network models~\cite{Guo:2017:ICML}.

\iffalse
is a visual representation of the calibration of NMT models. We partition the output tokens into several bins according to their prediction confidence. Then we calculate the average confidence and accuracy of each bin. We plot accuracy as a function of confidence, and calculate $\mathrm{gap} = |\mathrm{acc} - \mathrm{conf}|$ for each bin. Small gaps indicates well calibrated outputs.

Both the training ECE and the inference ECE are computed using the predictions of NMT models on the development. The difference is that when estimating training ECE, NMT models are fed with ground-truth prefixes~\cite{Kumar:2019:ICLR-Workshop,Muller:2019:arXiv} (i.e., next-token generation), while for inference ECE, NMT models are fed with machine translation prefixes. We find that the training ECE is very small, indicating that NMT models are well-calibrated in the scenario of next token prediction, confirming the findings of \citet{Kumar:2019:ICLR-Workshop,Muller:2019:arXiv}. However, the inference ECE is much higher, suggesting that NMT models suffer from miscalibration errors in inference. We also realize that NMT models are negatively affected by not only over-estimation but also under-estimation, different from classification neural network models~\cite{Guo:2017:ICML}.
% \zptu{try to expand and polish this paragraph based on the results.}
\fi

\subsection{Correlation with Translation Errors}

We investigated the calibration error of tokens with different TER labels. As the development set is small, to make the results more convincing, we sampled 100K sentences from the training set as a held-out set and retrained the NMT model on the remained training set excluding the held-out set. All results in this section is reported by the retrained model. %In other sections, the results are reported by models training on the whole training set if not explicitly stated. 
We firstly compute the gap between the confidence and the accuracy of each token in each confidence bin on the held-out set. Tokens in bins whose gaps are less than a threshold are labeled as well-calibrated, otherwise they are labeled as miscalibrated. We use the inference ECE estimated on the development set as the threshold for each language pair respectively. Miscalibrated tokens can be divided into two categories: over-estimation and under-estimation.

\begin{table}[t]
  \centering
  \begin{tabular}{c|c||c|c}
    \multicolumn{2}{c||}{\bf Translation}    &   \bf Well-Cali.   &   \bf Mis-Cali. \\
    \hline
    \hline
    \multirow{4}{*}{\rotatebox[origin=c]{90}{\bf Correct}} 
        & \em En-Jp & \bf 0.53  &   0.47 \\
        & \em En-De & \bf 0.57  &   0.43 \\
        & \em Zh-En & \bf 0.60  &   0.40 \\
        \cdashline{2-4}
        & All      & \bf 0.57  &   0.43 \\
    \hline
    \multirow{4}{*}{\rotatebox[origin=c]{90}{\bf Error}}
        & \em En-Jp & 0.46  &  \bf 0.54 \\
        & \em En-De & 0.43  &  \bf 0.57 \\
        & \em Zh-En & 0.36  &  \bf 0.63 \\
        \cdashline{2-4}
        & All      & 0.42  &  \bf 0.58 \\
  \end{tabular}
  \caption{Cosine similarity between the calibration and the translation errors on the held-out data. 
%   {\em Miscalibrated predictions correlate more to the translation errors.}
  }
  \label{tab:correlation-mistaken}
\end{table}

As shown in Table~\ref{tab:correlation-mistaken}, correct translations (i.e., ``C'') have higher correlations to well-calibrated predictions and erroneous translations (i.e., ``S'', ``I'', and ``D'') correlate more to miscalibrated predictions. This finding is more obvious when NMT models are trained on larger data (e.g., Zh-En).

Table~\ref{tab:correlation-error} lists the correlation between different translation errors and different kinds of miscalibration. We find that over-estimated predictions are closely correlated with over-translation and mis-translation errors, while the under-estimated predictions correlate well with under-translation errors. This finding demonstrates the necessity of studying inference calibration for NMT.

\begin{table}[t]
  \centering
  \begin{tabular}{c|c||cc}
    \multicolumn{2}{c||}{\bf Type}   &   \bf Under-Est.    &   \bf Over-Est.\\
    \hline
    \hline
    \multirow{4}{*}{\rotatebox[origin=c]{90}{\bf Under-Tra.}} 
        & En-Jp & \bf 0.35  &   0.22 \\
        & En-De & \bf 0.28  &   0.24 \\
        & Zh-En & 0.31  &   0.31 \\
        \cdashline{2-4}
        & All & \bf 0.32 & 0.26 \\
    \hline
    \multirow{4}{*}{\rotatebox[origin=c]{90}{\bf Over-Tra.}}
        & En-Jp & 0.28  &   \bf 0.32 \\
        & En-De & 0.20  &   \bf 0.36 \\
        & Zh-En & 0.29  &   \bf 0.35 \\
        \cdashline{2-4}
        & All & 0.26 & \bf 0.34 \\
    \hline
    \multirow{4}{*}{\rotatebox[origin=c]{90}{\bf Mis-Tra.}}
        & En-Jp & 0.24  &  \bf 0.36 \\
        & En-De & 0.17  &  \bf 0.42 \\
        & Zh-En & 0.24  &  \bf 0.40 \\
        \cdashline{2-4}
        & All & 0.21 & \bf 0.39 \\
  \end{tabular}
  \caption{Cosine similarity between the miscalibration errors (under-estimation and over-estimation) and the translation errors (under-translation, mis-translation, and over-translation) on the held-out data. 
%   {\em Over-estimated predictions correlate more with over-translation and mis-translation errors, while under-estimated predictions correlate more with under-translation errors.}
  }
  \label{tab:correlation-error}
\end{table}

\section{Linguistic Properties of Miscalibration}
\label{sec:lingual}

In this section, we investigate the linguistic properties of miscalibrated tokens in NMT outputs. 
We explore the following five types of properties: frequency, position, fertility, syntactic roles, and word granularity.
%~\footnote{Distributions on these properties are listed in Appendix.} 

Frequency is generally related to miscalibration; position, fertility, and word granularity are three factors associated with structured prediction; syntactic roles or linguistic roles may vary across language pairs. The results in this section are reported on the held-out set by the retrained model.

% \zptu{a paragraph summarize the following sections, try to find an example in my previous papers}

\paragraph{Relative Change} We use the relative change of the proportion of a certain category of tokens to quantify to what extent they suffer from the under/over-estimation. For instance, in the Zh-En task, high-frequency tokens account for 87.6\% on the whole held-out set, and among over-estimated tokens, high-frequency tokens account for 77.3\%, thus for over-estimation the relative change of high-frequency tokens is (77.3-87.6)/87.6=-11.76\% in Zh-En. Accordingly, the value of the red rectangle of Zh-En is -11.76\% in Figure~\ref{fig:freq-over}. 

Positive relative change denotes that a certain type of linguistic property accounts more in miscalibrated predictions than in all the predictions, suggesting this type of linguistic property suffers from the miscalibration problem. Similarly, negative relative change suggests that a certainty type of linguistic property is less likely to be impaired by the miscalibration problem. 

% \zptu{in the following sections, first paragraph: the setting; followed by the observations.}

% \zptu{re-summarize the findings, from the perspectives of miscalibration instead of from the linguistic properties as listed below}

\subsection{Frequency}

\begin{figure}[t]
    \centering
    % \vspace{-10pt}
    \subfloat[Over-Estimation]{
    \label{fig:freq-over}
    \includegraphics[height=0.211\textwidth]{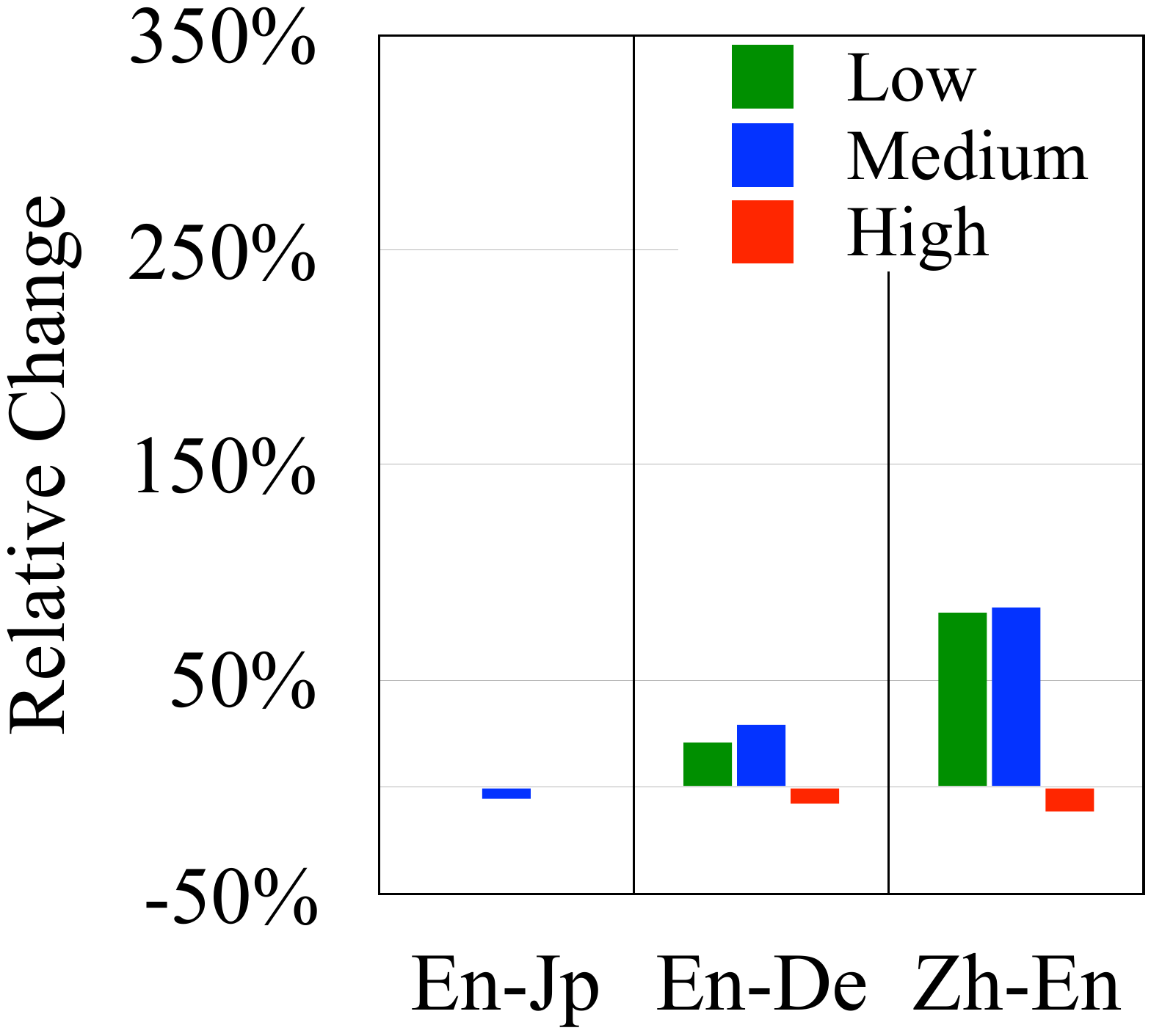}}
    \hspace{0.02\textwidth}
    \subfloat[Under-Estimation]{
    \label{fig:freq-under}
    \includegraphics[height=0.211\textwidth]{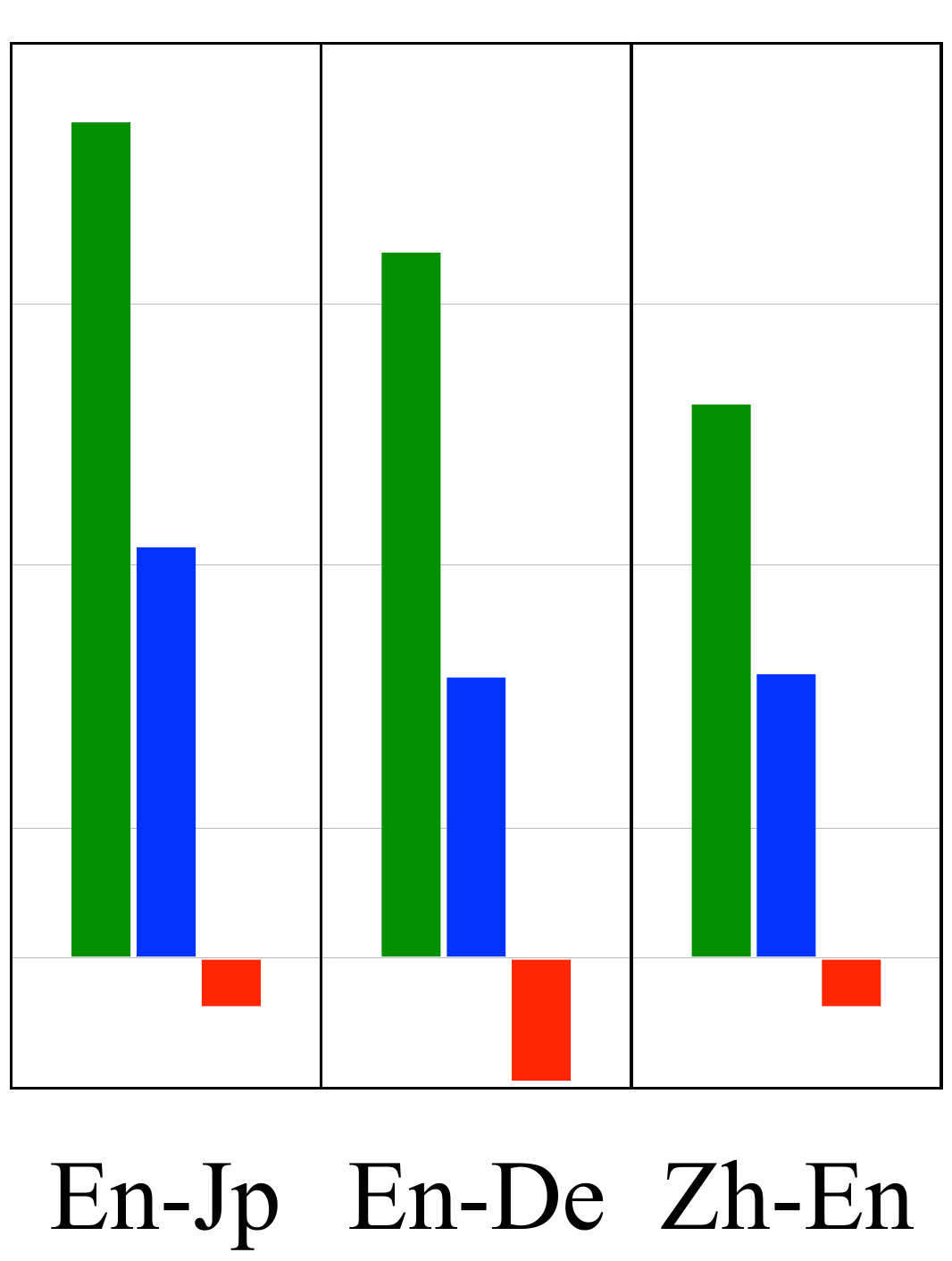}}
\caption{Effect of frequency on miscalibration.}
\label{fig:frequency}
\end{figure}

We divide tokens into three categories based on their frequency, including {\em High}: the most 3,000 frequent tokens; {\em Medium}: the most 3,001-12,000 frequent tokens; {\em Low}: the other tokens.

{\em Low-frequency tokens are miscalibrated in the direction of under-estimation.} As shown in Figure~\ref{fig:frequency}, 
the relative changes of low- and medium-frequency tokens are much bigger than those of high-frequency tokens. The under-estimation in low- and medium-frequency tokens can be alleviated by increasing the size of training data (Figure~\ref{fig:freq-under}, data size: En-Jp $<$ En-De $<$ Zh-En).

{\em Low-frequency tokens contribute more to over-estimation.} As shown in Figure~\ref{fig:freq-over}, the relative changes of low- and medium-frequency tokens are positive while those of high-frequency tokens are negative, regarding over-estimation.

{\em High-frequency tokens are less likely to be miscalibrated.} We find the relative changes of high frequency tokens are negative across the three language pairs. The imbalance in token frequency plays an important role in the calibration of NMT.

\subsection{Position}

\iffalse
\begin{table}[h]
  \centering
  \begin{tabular}{c|rrr}
    {\bf Position}  & \bf En-Jp    &   \bf En-De   &   \bf Zh-En \\
    \hline
    Left & 34.4\% & 34.5\% & 34.6\%\\
    Middle & 33.3\% & 33.3\% & 33.3\%\\
    Right & 32.2\% & 32.1\% & 32.0\%\\
  \end{tabular}
  \caption{Distribution of position.}
  \label{tab:position-distribution}
\end{table}
\fi

% \zptu{setting, motivation, hypothesis if exists}

\begin{figure}[t]
    \centering
    \subfloat[Over-Estimation]{
    \label{fig:position-over}
    \includegraphics[height=0.211\textwidth]{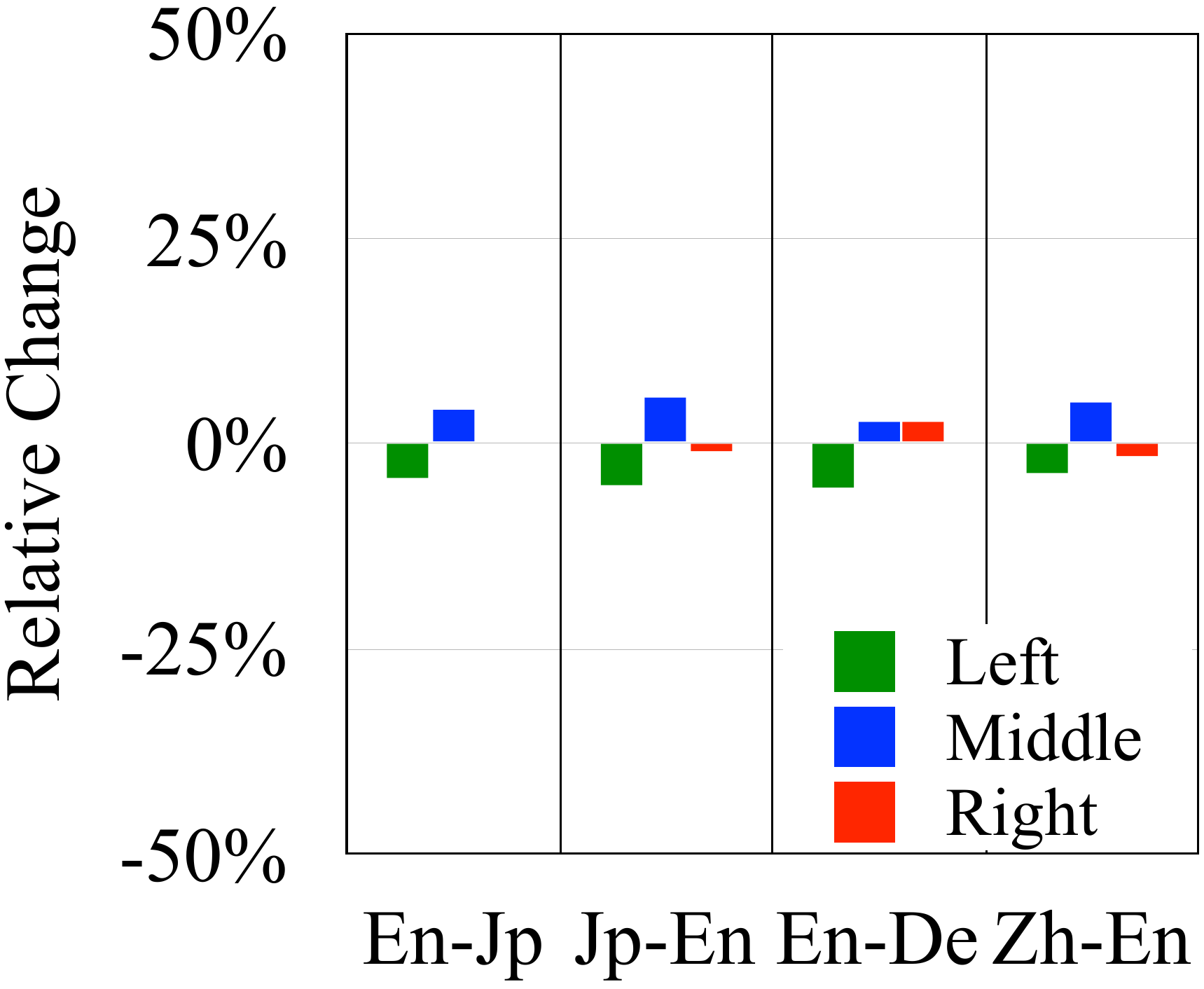}}
    \hspace{0.02\textwidth}
    \subfloat[Under-Estimation]{
    \label{fig:position-under}
    \includegraphics[height=0.211\textwidth]{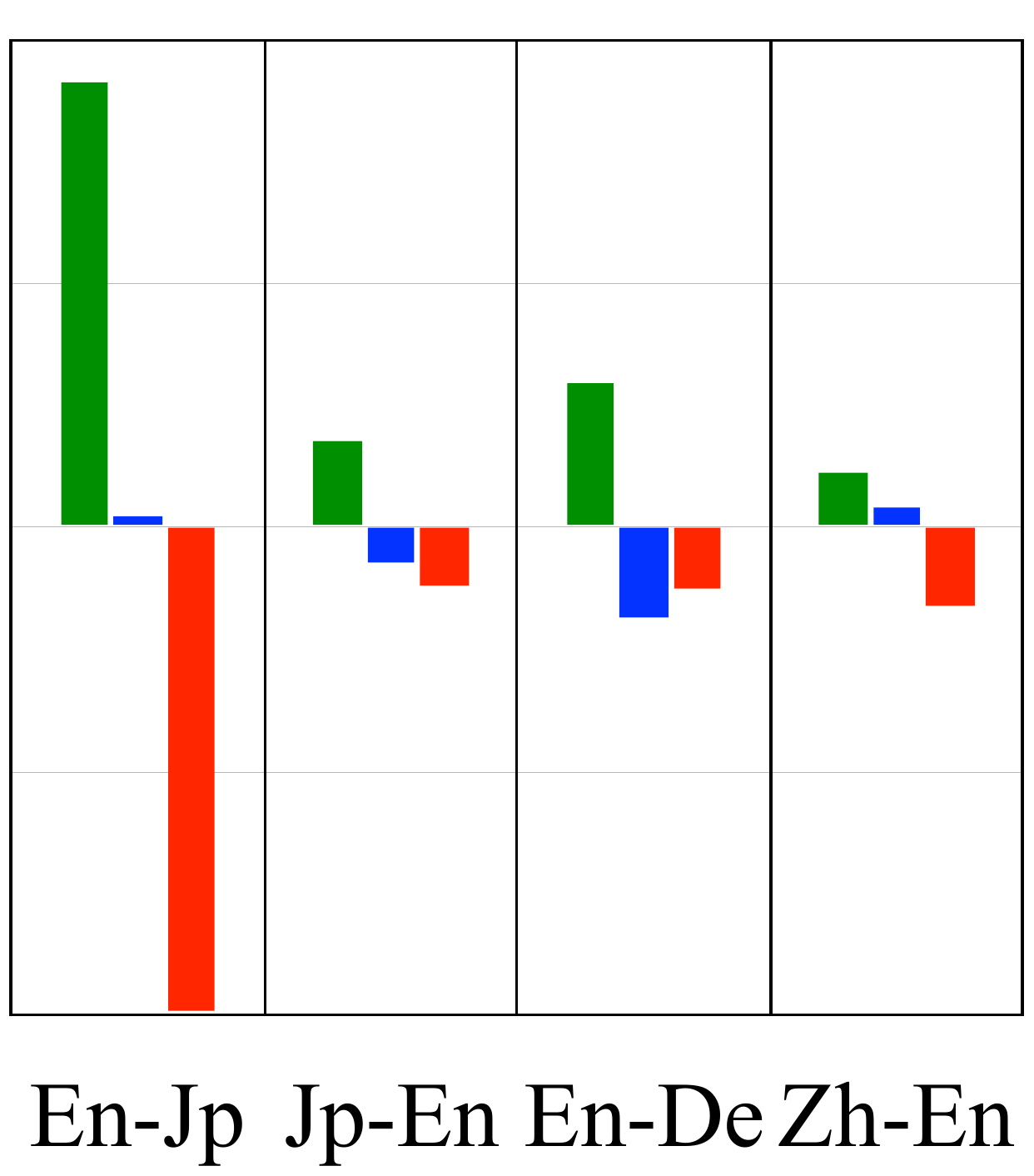}}
\caption{Effect of relative position on miscalibration.}
\label{fig:position}
\end{figure}

In structured prediction, different positions may behave differently regarding miscalibration. Thus we divide all the tokens equally into three categories: {\em Left}: tokens on the left third; {\em Middle}: tokens on the middle third; {\em Right}: tokens on the right third.
% \begin{itemize}
%     \item {\em Left}: tokens on the left third;
%     \item {\em Middle}: tokens on the middle third;
%     \item {\em Right}: tokens on th right third.
% \end{itemize}
Figure~\ref{fig:position} depicts the relative changes of these three positions. Since Japanese is a head-final language~\cite{Wu:2018:EMNLP}, we also include the results of Japanese-English (``Jp-En'') for comparison.

{\em Over-estimation does not have a bias on position.} And this holds for both left-branching and right-branching languages. Increasing the size of training data is less likely to affect the over-estimation in different positions.

{\em Under-estimation occurs more in the left part.} This phenomenon is more obvious in left-branching languages (e.g., Japanese) than in right-branching languages (e.g., English and German), confirming that characteristics of a language play an important role in machine translation~\cite{Wu:2018:EMNLP}.

\subsection{Fertility}

\begin{figure}[h]
    \centering
    \subfloat[Over-Estimation]{
    \includegraphics[height=0.209\textwidth]{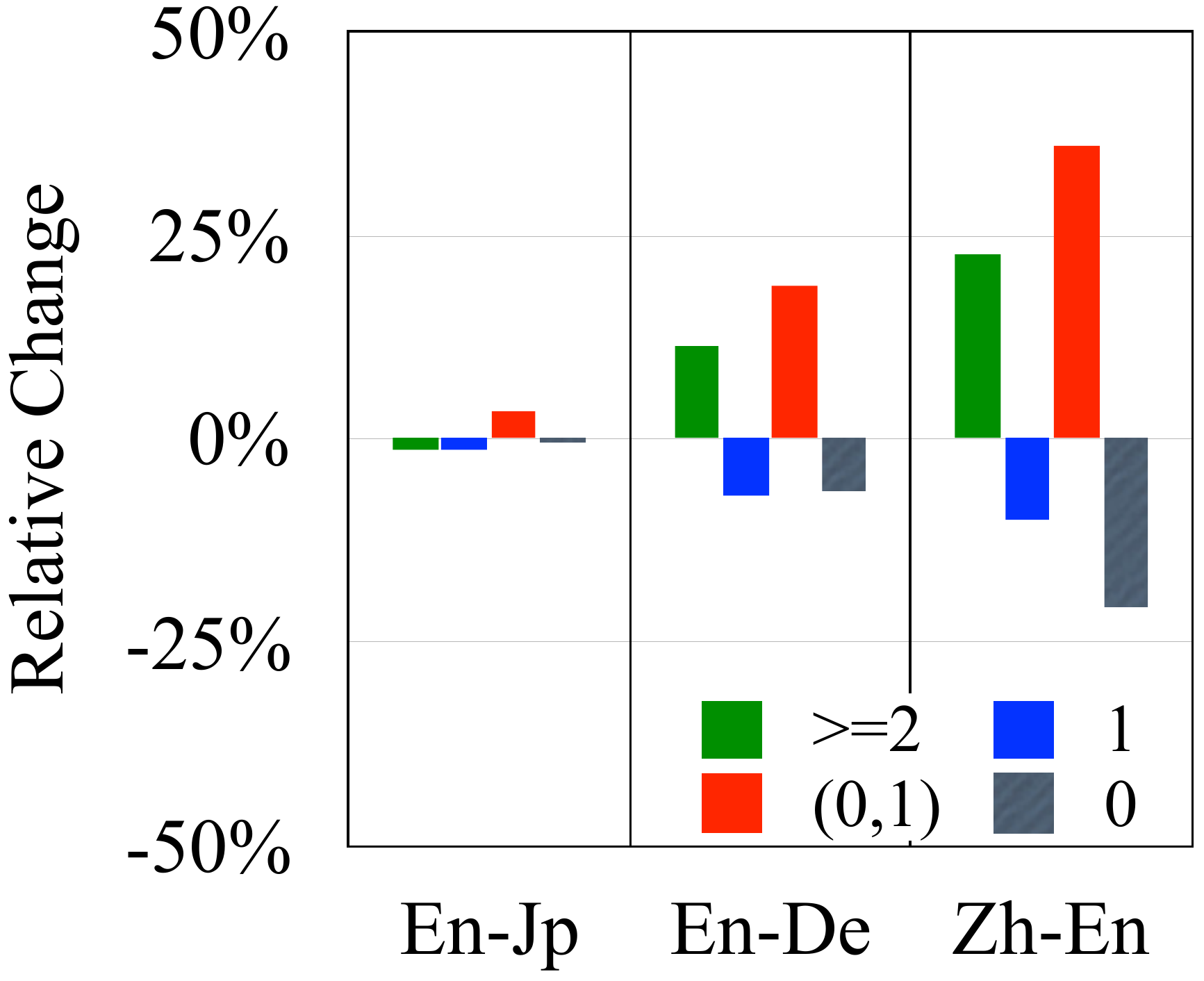}}
    \hspace{0.02\textwidth}
    \subfloat[Under-Estimation]{
    \includegraphics[height=0.209\textwidth]{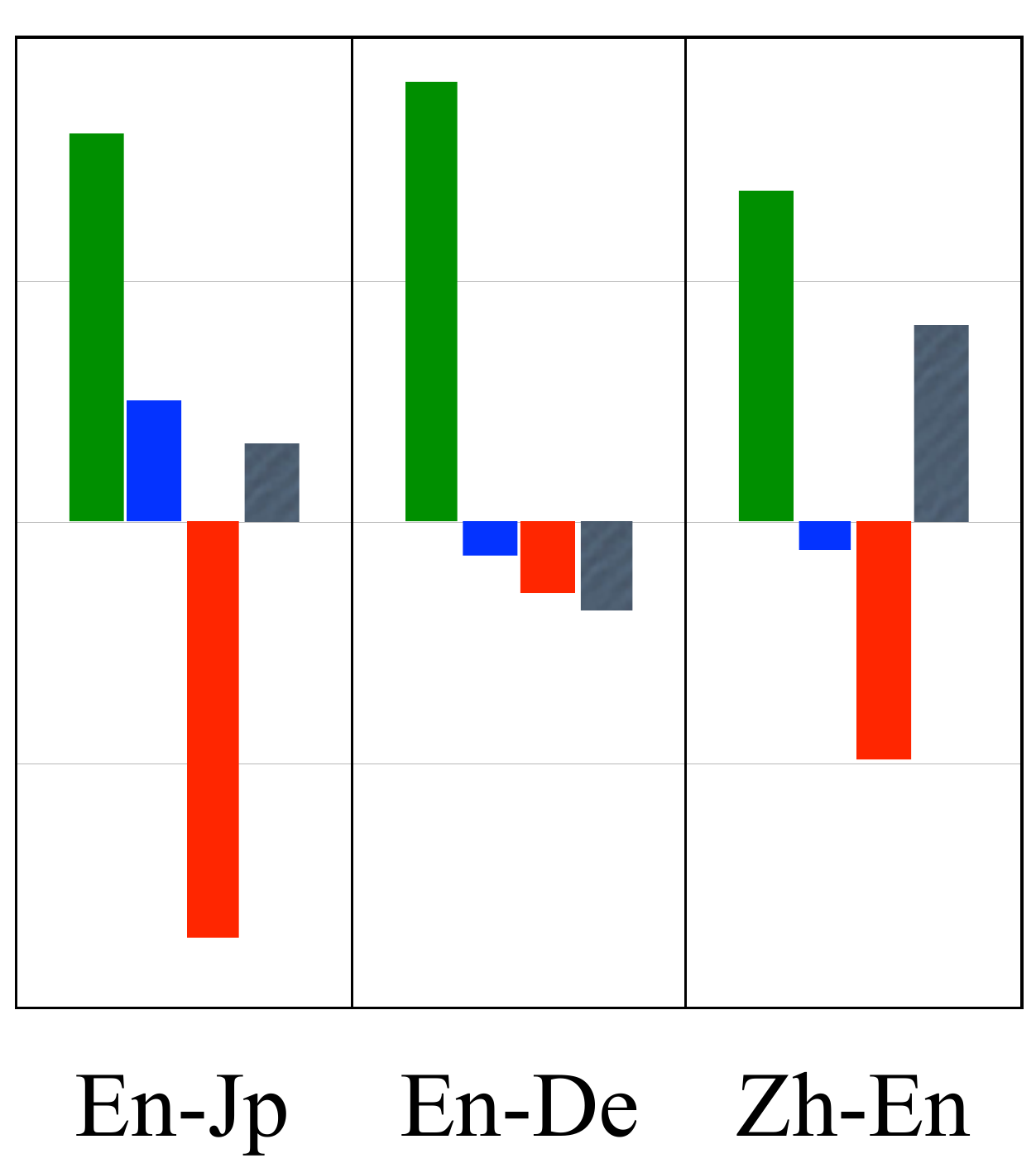}}
\caption{Effect of fertility on miscalibration.}
\label{fig:fertility}
\end{figure}

\begin{figure*}[h]
    \centering
    \subfloat[Over-Estimation]{
    \includegraphics[height=0.211\textwidth]{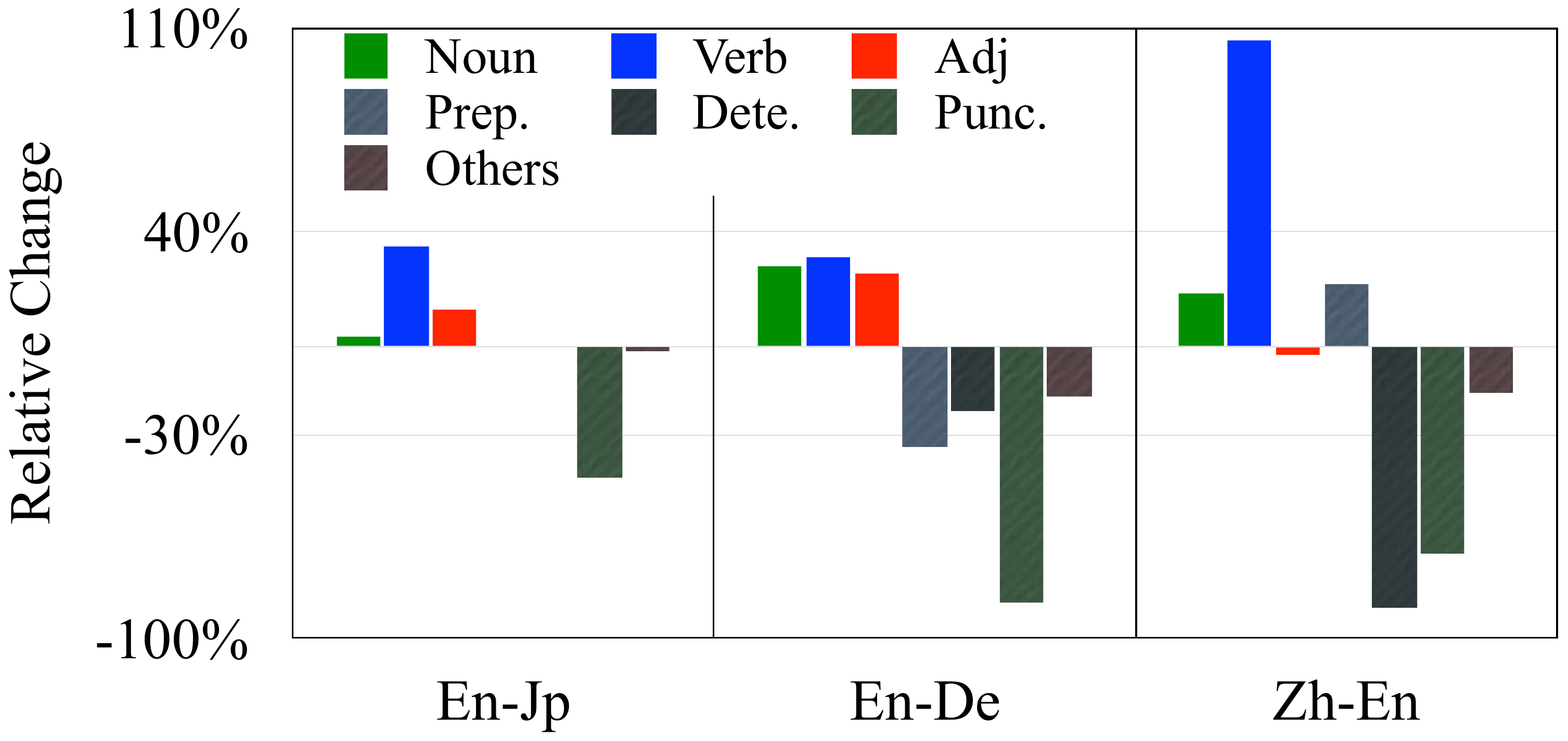}}
    \hspace{0.05\textwidth}
    \subfloat[Under-Estimation]{
    \includegraphics[height=0.211\textwidth]{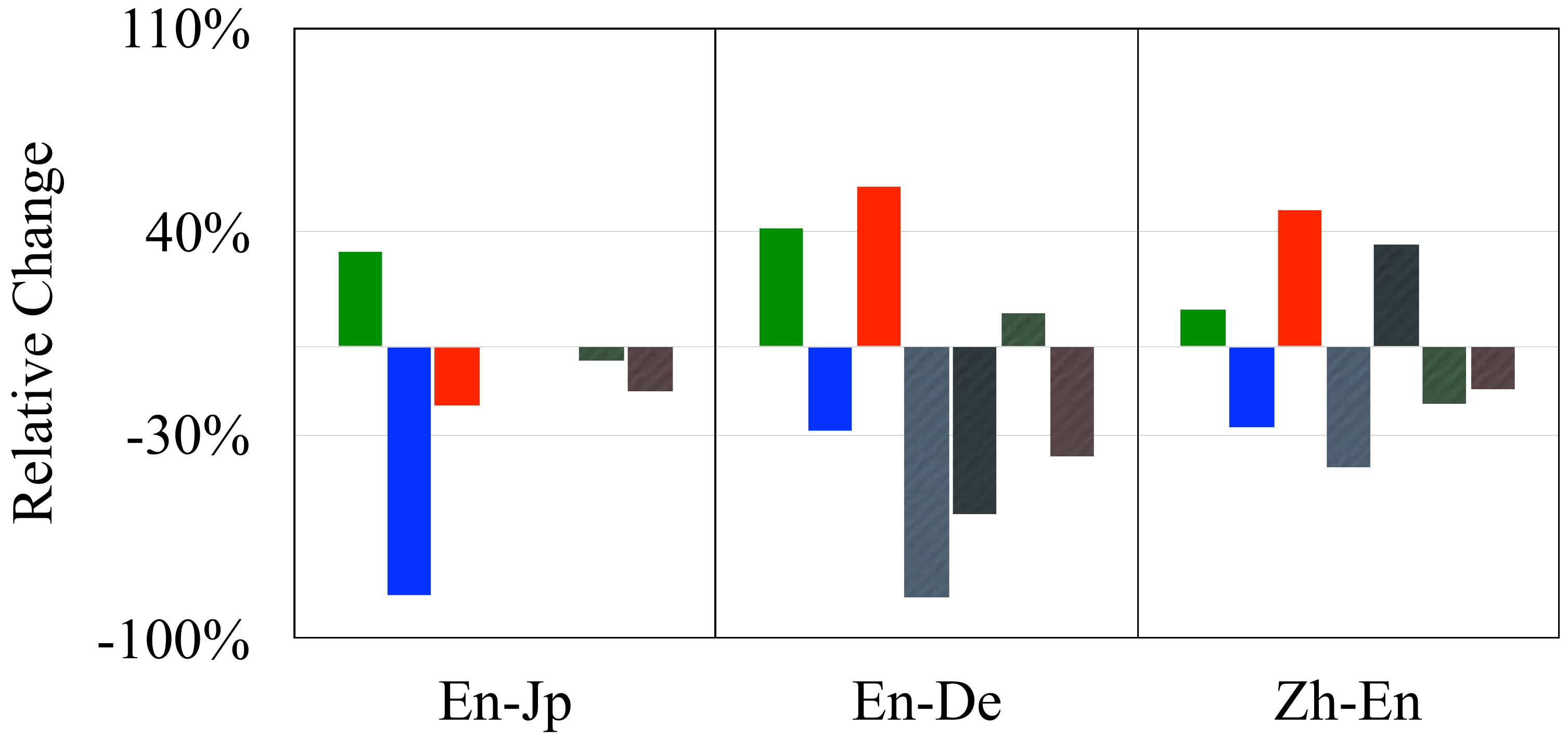}}
\caption{Effect of POS tags on miscalibration.}
\label{fig:pos-tags}
\end{figure*}

Fertility indicates how many source tokens a target token is aligned to, which is highly related to  inference in NMT.
We use \verb|Fast Align|~\cite{Dyer:2013:NAACL} to extract bilingual alignment. 
We distinguish between four categories regarding fertility: ``{\bf $\ge2$}'': target tokens that are aligned to more than one source tokens; ``{\bf $1$}'': target tokens that are aligned to a single source token; ``{\bf $(0, 1)$}'': target tokens that are aligned to a single source token along with other target tokens; ``{\bf $0$}'': target tokens that are not aligned to any source token. 
Figure~\ref{fig:fertility} plots the results.
% relative change of different fertility properties.

% \zptu{setting, motivation, hypothesis if exists}

{\em Tokens aligning to less than one source token suffer from over-estimation.} The extent grows with the data size. In addition, these tokens (``(0, 1)'') are less likely to suffer from under-estimation.

{\em Tokens aligning to more than one source token suffer more from under-estimation.} The relative change of “fertility$>=$2” is much larger than that of the other types of fertility. Meanwhile, the null-aligned target tokens (“fertility$=$0”) also suffer from under-estimation problem instead of over-estimation problem on the large-scale Zh-En data.

\subsection{Syntactic Roles}

In this experiment, we investigate the syntactic roles of miscalibrated tokens.~\footnote{If a token is a sub-word segmented by BPE, the token shares the syntactic role of the full word that it belongs to.} 
Words in English and German sentences are labeled by \verb|Stanford POS tagger|\footnote{https://nlp.stanford.edu/software/tagger.shtml}, 
and Japanese sentences are labeled by \verb|Kytea|\footnote{http://www.phontron.com/kytea/}.
We distinguish between the following POS tags: noun, verb, adjective, preposition, determiner, punctuation, and the others. 
Noun, verb, and adjective belong to content tokens. Preposition, determiner, punctuation and the others belong to content-free tokens.

{\em Content tokens are more likely to suffer from miscalibration.} From Figure~\ref{fig:pos-tags} we find that the most relative changes of content tokens (i.e., ``Noun'', ``Verb'' and ``Adj'') are positive, while most of the relative changes of the content-free tokens (i.e., ``Prep.'', ``Dete.'', ``Punc.'', ``Others'') are negative. Among content tokens, the verbs (``Verb'') face the over-estimation problem instead of the under-estimation problem. Surprisingly, the adjectives (``Adj'') suffer from under-estimation problem on large data (e.g., En-De and Zh-En).

\begin{figure}[t]
    \centering
    \subfloat[Over-Estimation]{
    \label{fig:bpe-over}
    \includegraphics[height=0.211\textwidth]{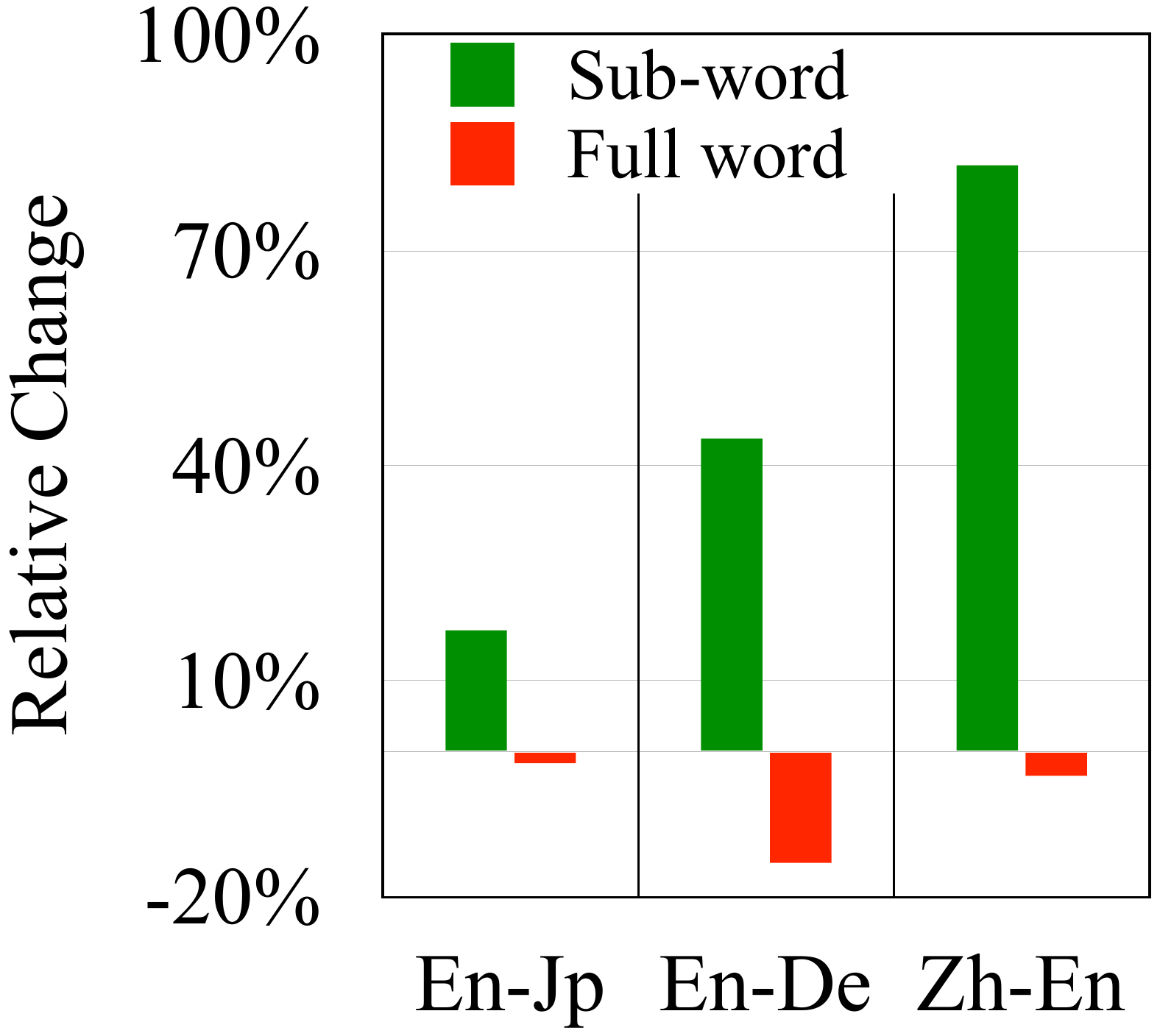}}
    \hspace{0.02\textwidth}
    \subfloat[Under-Estimation]{
    \label{fig:bpe-under}
    \includegraphics[height=0.211\textwidth]{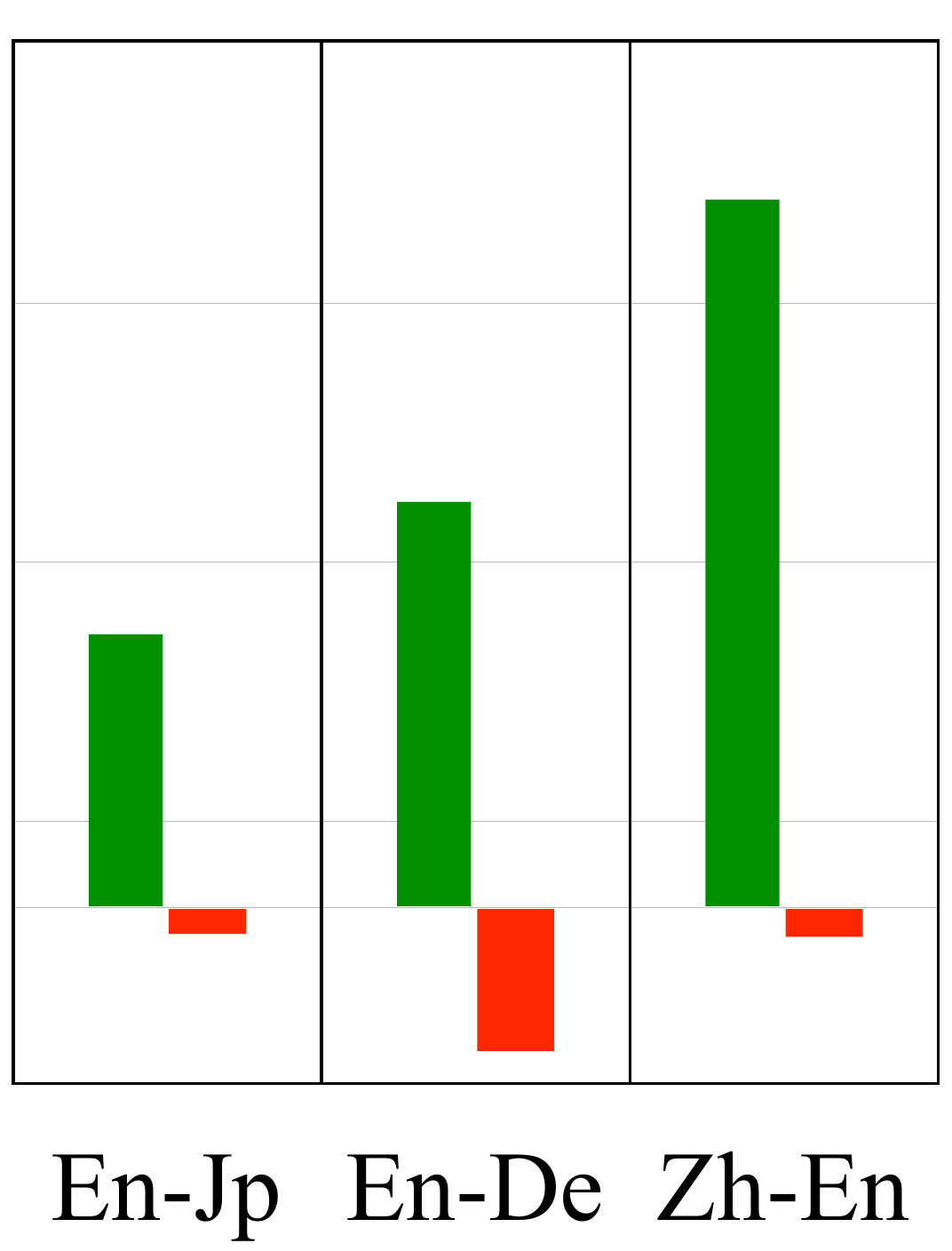}}
\caption{Effect of word granularity on miscalibration.}
\label{fig:bpe}
\end{figure}

\subsection{{Word Granularity}}
BPE segmentation is the preliminary step for current NMT systems, which may segment some words into sub-words. To explore the effect of word granularity on the miscalibration of NMT models, we divide the tokens after BPE segmentation into two categories: {\em Sub-Words} that are divided into word fragments by BPE (e.g., with ``@@''), and {\em Full Words} that are not divided by BPE. Figure~\ref{fig:bpe} depicts the results.

{\em Sub-words suffer more from miscalibration, while full words are less likely to be miscalibrated.} The relative changes of sub-words are all positive for both over- and under-estimation, while those of full words are all negative. 
~\newcite{Sennrich:2016:ACL} showed that BPE addresses the open-vocabulary translation by encoding rare and unknown words as sequences of sub-word units. Our results confirm their claim: the behaviors of sub-words and full words correlate well with those of low- and high-frequency tokens respectively.

\begin{table*}[t]
  \centering
  \begin{tabular}{c|c||c|rrr||c|rrr}
    \bf Label   &  \multirow{2}{*}{\bf Dropout} & \multicolumn{4}{c||}{\bf Beam Size = 10}  &   \multicolumn{4}{c}{\bf Beam Size = 100}  \\
    \cline{3-10}
    \bf Smoothing & & \em BLEU & \em ECE & \em Over. & \em Under. & \em BLEU & \em ECE & \em Over. & \em Under. \\
    \hline
    \texttimes & \texttimes & 23.03 & 25.49 & 58.3\% & 9.6\% 
                            & 22.90 & 26.46 & 59.4\% & 9.3\% \\
    \hdashline
    \checkmark & \texttimes & 24.51 & 14.99 & 42.3\% & 17.3\% 
                            & 24.58 & 15.97 & 42.8\% & 16.9\% \\
    \texttimes & \checkmark & 27.52 & 20.75 & 52.3\% & 10.1\% 
                            & 26.93 & 22.57 & 53.6\% & 9.8\%  \\
    \hdashline
    \checkmark & \checkmark & 27.65 & 14.26 & 39.7\% & 14.1\% 
                            & 27.68 & 14.75 & 40.1\% & 14.2\%  \\
    \hline
    \textproc{Graduated} & \checkmark & \bf 27.76 & \bf 5.07 & 29.1\% & 31.6\% 
                              & \bf 28.07 & \bf 5.23 & 29.5\% & 31.4\%
  \end{tabular}
  \caption{Results of label smoothing and dropout on the En-De task. ``Over.'' and ``Under.'' denote over-estimation and under-estimation, respectively.}
  \label{tab:ls-dropout}
\end{table*}

\begin{figure*}[ht]
    \centering
    \subfloat[None]{
    \includegraphics[height=0.24\textwidth]{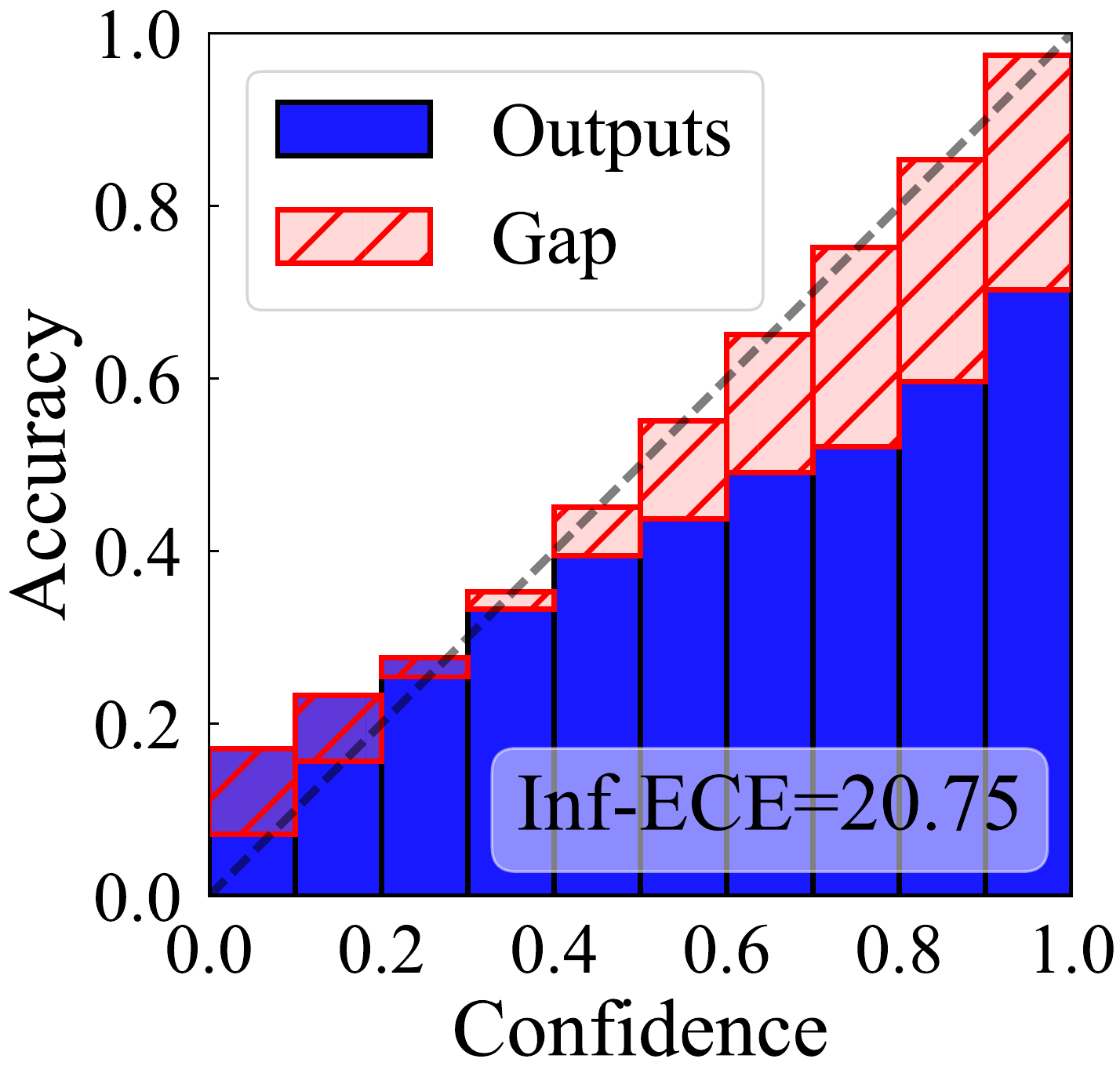} 
    } 
    \hspace{0.05\textwidth}
    \subfloat[Vanilla]{
    \includegraphics[height=0.24\textwidth]{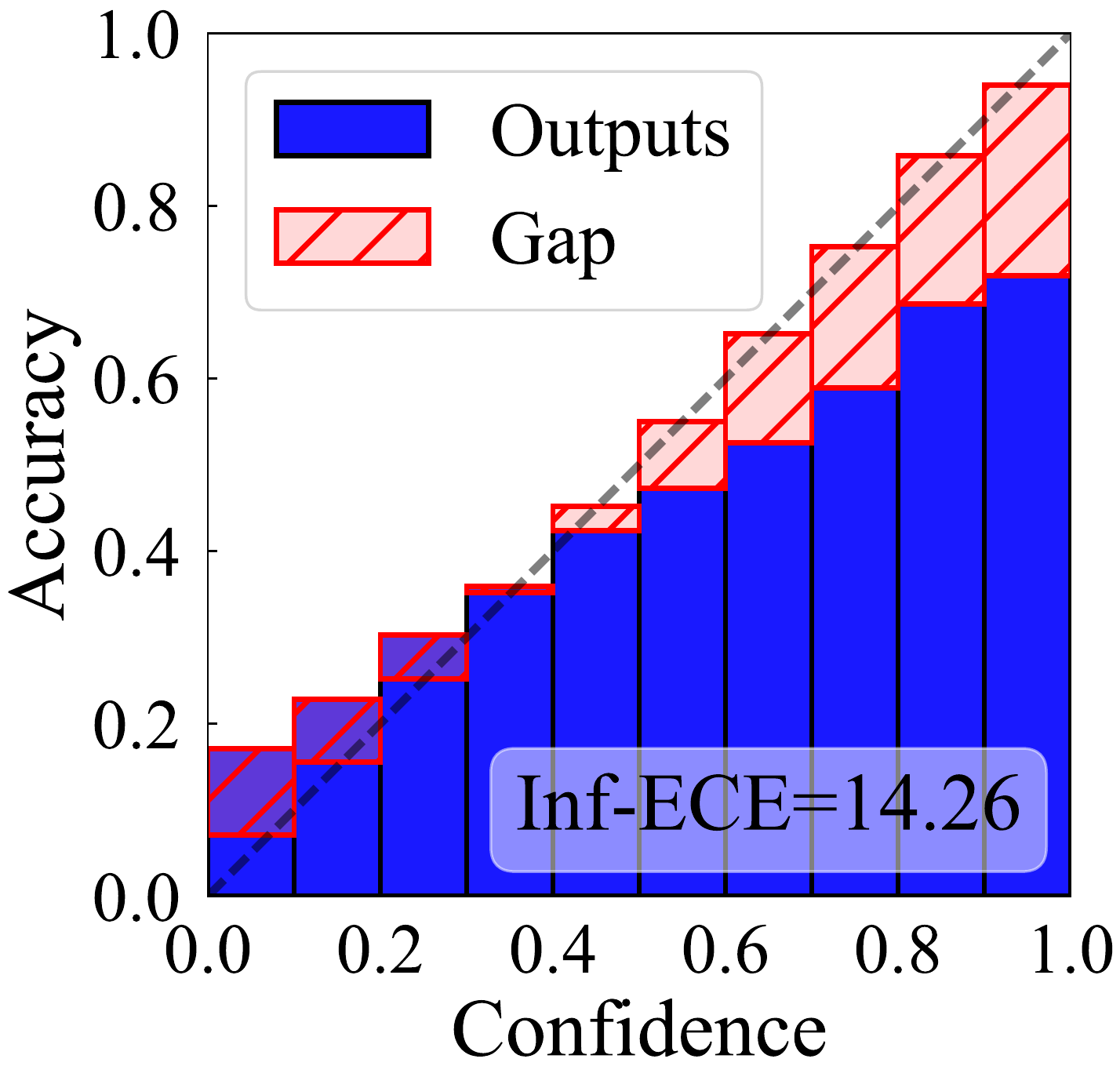} 
    }
    \hspace{0.05\textwidth}
    \subfloat[Graduated]{
    \includegraphics[height=0.24\textwidth]{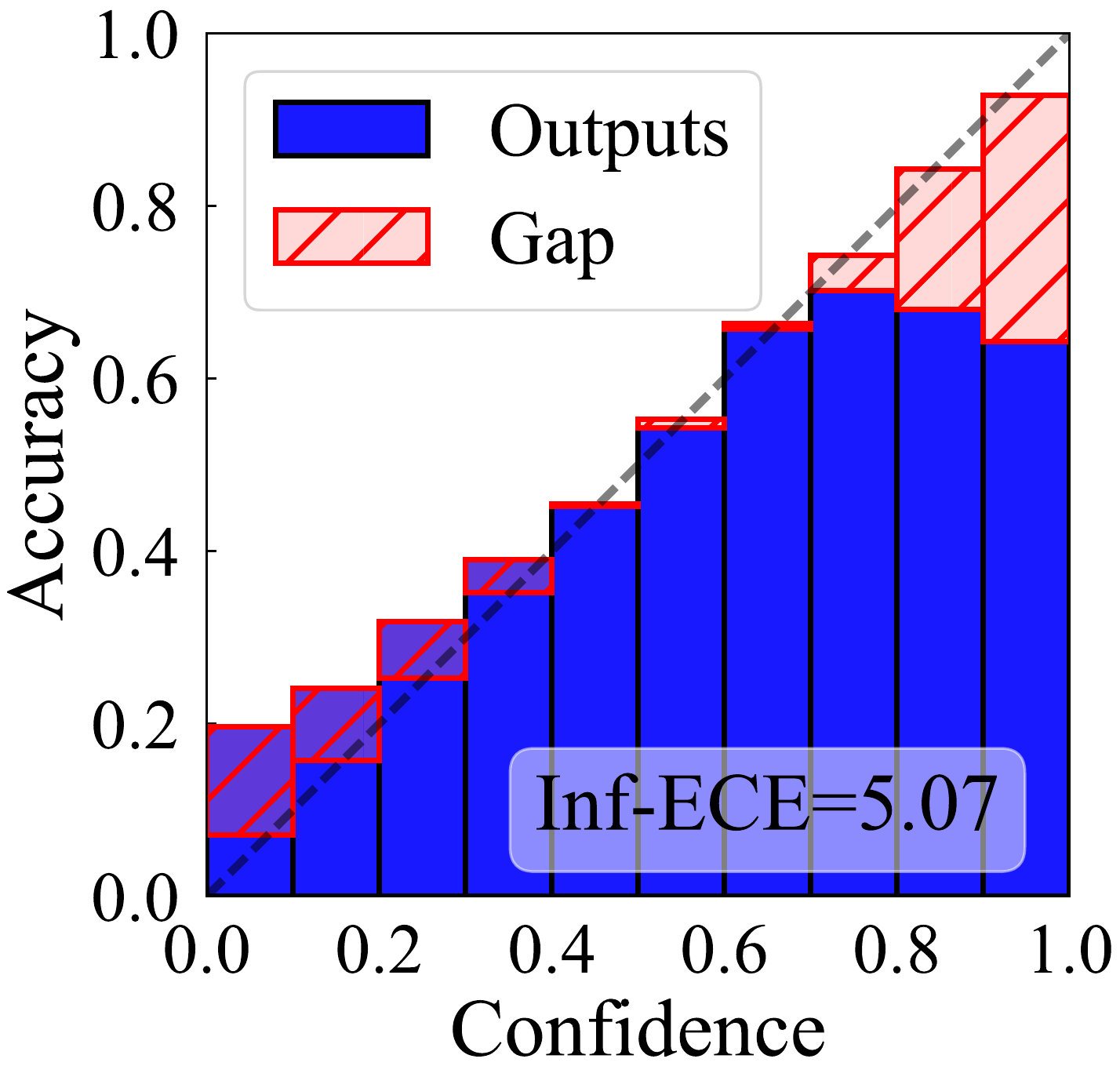} 
    }
\caption{Reliability diagrams of different label smoothing strategies: (a) no label smoothing; (b) vanilla label smoothing; (c) graduated label smoothing. The results are reported on the WMT14 En-De translation task.}
\label{fig:tech_smoothing}
\end{figure*}

\section{Revisiting Advances in Architecture and Regularization}

\citet{Guo:2017:ICML} have revealed that the miscalibration on classification tasks is closely related to lack of regularization and increased model size. In this section we check whether the conclusion holds on the inference of NMT models, which belong to a family of structured generation.

One criticism of NMT inference is that the translation performance inversely decreases with the increase of search space~\cite{Tu:2017:AAAI}. Quite recently,~\newcite{Kumar:2019:ICLR-Workshop} claimed that this problem can be attributed to miscalibration. Accordingly, we also report results on large beam size and find that reducing miscalibration can improve the NMT performance in large beam size.

\begin{table*}[t]
  \centering
  \begin{tabular}{c|c|r||c|rrr||c|rrr}
    \multirow{2}{*}{\bf Enc.} & \multirow{2}{*}{\bf Dec.} &  \multirow{2}{*}{\bf Para.}  & \multicolumn{4}{c||}{\bf Beam Size = 10}  &  \multicolumn{4}{c}{\bf Beam Size = 100} \\
   \cline{4-11}
    & & & \em BLEU & \em ECE & \em Over. & \em Under. 
          & \em BLEU & \em ECE & \em Over. & \em Under. \\
    \hline
    \textproc{base} & \textproc{base} & 88M & 27.65 &  14.26 & 39.7\% & 14.1\% & 27.68 & 14.75 & 40.1\% & 14.2\% \\
    \hline \hline
    \textproc{deep} & \textproc{deep} & 220M & 28.86 & 14.99 & 40.3\% & 14.1\% & 28.64 & 15.55 & 41.8\% & 14.0\% \\
    \hdashline
    \textproc{deep} & \textproc{base} & 145M & \bf 29.09 & \bf 14.28 & 39.6\% & 14.1\% & \bf 29.29 & \bf 14.53 & 39.6\% & 14.2\% \\
    \hline \hline
    \textproc{wide} & \textproc{wide} & 264M & 28.66 & 16.09 & 42.3\% & 12.6\% & 28.42 & 17.22 & 43.2\% & 12.5\% \\
    \hdashline
    \textproc{wide} & \textproc{base} & 160M & \bf 28.97 & \bf 14.83 & 39.7\% & 13.6\% & \bf 29.09 & \bf 15.06 & 39.8\% & 13.7\% \\
  \end{tabular}
  \caption{Effect of model size by enlarging encoder (``Enc.'') and decoder (``Dec.'') on the En-De dataset.}
  \label{tab:capacity-architecture}
\end{table*}

\subsection{Regularization Techniques}
\label{sec:training}

We revisit two important regularization techniques that directly affect confidence estimation:

\begin{itemize}
    \item {\em Label Smoothing}~\cite{Szegedy:2016:CVPR}: distributing a certain percentage of confidence from the ground truth label to other labels uniformly in training.
    \item{\em Dropout}~\cite{Hinton:2012:arXiv}: randomly omitting a certain percentage of the neural networks on each training case, which has been shown effective to prevent the over-fitting problem for large neural networks.
\end{itemize}

\iffalse
We believe that label smoothing on structured prediction with large vocabulary (e.g., machine translation) can be further improved by solving the following two potential constraints:
\begin{itemize}
    \item It assigns a unified smoothing parameter $\epsilon$ to all generated tokens, regardless of the distinct properties of the tokens. For example, a miscalibrated prediction should be assigned a higher $\epsilon$ than a well-calibrated prediction.
    \item It only penalizes the groundtruth label, while ignore the incorrect labels during the training phase.
    Recent studies show that overlooking incorrect labels leads to the representation degeneration problem: most of the learnt word embeddings tend to degenerate and be distributed into a narrow cone~\cite{Gao:2019:ICLR}.
\end{itemize}
\fi

For comparison, we disable label smoothing or dropout to retrain the model on the whole training set. The results are shown in Table~\ref{tab:ls-dropout}. We find that label smoothing improves the performance by greatly reducing the over-estimation, at the cost of increasing the percentage of under-estimation error. Dropout alleviates the over-estimation problem, and does not aggravate under-estimation. Although label smoothing only marginally improves performance on top of dropout, it is essential for maintaining the translation performance in larger search space (i.e., Beam Size = 100).

As seen from Table~\ref{tab:ls-dropout}, reducing ECE can only lead to marginal BLEU gains. We attribute this phenomenon to the fact that ECE is another metric to evaluate NMT models, which is potentially complementary to BLEU. Accordingly, ECE is not necessarily strictly negatively related to BLEU.

\paragraph{Graduated Label Smoothing}
Inspired by this finding, we propose a novel {\em graduated label smoothing} approach, in which the smoothing penalty for high-confidence predictions is bigger than that for low-confidence predictions. We firstly use the model trained by vanilla label smoothing to estimate the confidence of each token in the training set, then we set the smoothing penalty to 0.3 for tokens with confidence above 0.7, 0.0 for tokens with confidence below 0.3, and 0.1 for the remaining tokens. 

As shown in Table~\ref{tab:ls-dropout}, the graduated label smoothing can improve translation performance by alleviating inference miscalibration, and the improvement is more significant in large beam size. 
Figure~\ref{fig:tech_smoothing} shows the reliability diagrams of different label smoothing strategies. 
The graduated label smoothing can effectively calibrate the predictions with $0.4\leq$ confidence $\leq0.8$, while is less effective for low- (i.e., $<0.4$) and high-confidence (i.e., $>0.8$) predictions. We believe that the design of more advanced techniques to solve this problem is a worthwhile future direction of research.

\subsection{Increased Model Size}

The model size of NMT models has increased significantly recently~\cite{Bahdanau:2015:ICLR,Vaswani:2017:NeurIPS,Wang:2019:ACL}. We evaluated the inference calibration of models with different sizes. 
%The Transformer-base consists of an encoder and an decoder, each contains 6 layers with the hidden size of 512. 
We increase model size in the following two ways:
\begin{itemize}
    \item {\em Deeper model}: both the encoder and the decoder are deepened to 24 layers;
    \item {\em Wider model}: the hidden size of the encoder and the decoder is widened to 1024.
\end{itemize}
The BLEU score and inference ECE of different models are shown in Table~\ref{tab:capacity-architecture}.

{\em Increasing model size negatively affects inference calibration.}
We find that increasing both the encoder and the decoder increases the inference calibration error despite increasing the BLEU, confirming the finding of \citet{Guo:2017:ICML} that increased model size is closely related to model miscalibration. This leads to a performance drop in a larger search space (i.e., Beam Size = 100).

{\em Only enlarging the encoder improves translation quality while maintaining inference calibration.}
As the decoder is more directly related to the generation, it is more likely to result in miscalibration. In order to maintain the performance improvement and do not aggravate over-estimation, we propose to only increase the size of encoder and keep the decoder unchanged. Results in Table~\ref{tab:capacity-architecture} indicate that only enlarging the encoder can achieve better performance with fewer parameters compared to enlarging both the encoder and the decoder. In a larger search space (i.e., Beam Size = 100), models with high inference ECE will generate worse translations while models with low inference ECE can achieve improved translation performance.

\section{Conclusion}

Although NMT models are well-calibrated in training, we observe that they still suffer from miscalibration during inference because of the discrepancy between training and inference. 
Through a series of in-depth analyses, we report several interesting findings which may help to analyze, understand and improve NMT models. 
We revisit recent advances and find that label smoothing and dropout play key roles in calibrating modern NMT models. We further propose graduated label smoothing that can reduce the inference calibration error effectively.
% , making NMT models achieve better performance in large search space. 
Finally, we find that increasing model size can negatively affect the calibration of NMT models and this can be alleviated by only enlarging the encoder. As well-calibrated confidence estimation is more likely to establish trustworthiness with users, we plan to apply our work to interactive machine translation scenarios in the future.

\section*{Acknowledgments}
We thank all anonymous reviewers for their valuable comments and suggestions for this work.
This work was supported by the National Key R\&D Program of China (No. 2017YFB0202204), National Natural Science Foundation of China (No. 61925601, No. 61761166008, No. 61772302), Beijing Advanced Innovation Center for Language Resources (No. TYR17002),  and the NExT++ project supported by the National Research Foundation, Prime Ministers Office, Singapore under its IRC@Singapore Funding Initiative.

\bibliographystyle{acl_natbib}
\bibliography{main.bib} 
\end{document}